\documentclass[10pt,twocolumn,letterpaper]{article}

\usepackage[pagenumbers]{iccv} 
\usepackage{algorithm}
\usepackage{algpseudocode}
\usepackage{float}

\newcommand{\set}[1]{\mathcal{#1}}

\definecolor{iccvblue}{rgb}{0.21,0.49,0.74}
\usepackage[pagebackref,breaklinks,colorlinks,allcolors=iccvblue]{hyperref}

\def\paperID{8650} 
\def\confName{ICCV}
\def\confYear{2025}

\title{Multi-View Industrial Anomaly Detection with Epipolar Constrained Cross-View Fusion}
\author{
    Yifan Liu$^{1,2}$ \quad
    Xun Xu$^{1}$ \quad
    Shijie Li$^{1}$ \quad
    Jingyi Liao$^{1}$ \quad
    Xulei Yang$^{1}$ \\
    $^1$ Institute for Infocomm Research (I$^2$R), A*STAR, Singapore \\
    $^2$ National University of Singapore \\
}

\begin{document}
\maketitle

\begin{abstract}
Multi-camera systems provide richer contextual information for industrial anomaly detection. However, traditional methods process each view independently, disregarding the complementary information across viewpoints. Existing multi-view anomaly detection approaches typically employ data-driven cross-view attention for feature fusion but fail to leverage the unique geometric properties of multi-camera setups. In this work, we introduce an epipolar geometry-constrained attention module to guide cross-view fusion, ensuring more effective information aggregation. To further enhance the potential of cross-view attention, we propose a pretraining strategy inspired by memory bank-based anomaly detection. This approach encourages normal feature representations to form multiple local clusters and incorporate multi-view aware negative sample synthesis to regularize pretraining. We demonstrate that our epipolar guided multi-view anomaly detection framework outperforms existing methods on the state-of-the-art multi-view anomaly detection dataset.

\end{abstract}    
\section{Introduction}
\label{sec:intro}

Detecting defects is crucial for quality assurance in industrial processes~\cite{liu2024deep, xie2024iad}. Due to the diverse and unconstrained nature of defects, this task is often formulated as unsupervised anomaly detection. The advancement of industrial anomaly detection has been largely driven by the introduction of diverse and realistic datasets~\cite{bergmann2019mvtec,zou2022spot}. However, most existing methods focus on analyzing defects from a single snapshot of an object. In practical industrial settings, production lines are often equipped with multiple cameras, capturing objects from different viewpoints. This setup has led to the emergence of studies in multi-view industrial anomaly detection (MVIAD)~\cite{wang2024real}.  
\begin{figure}[!h]
  \centering
    \centering
    \includegraphics[width=\linewidth]{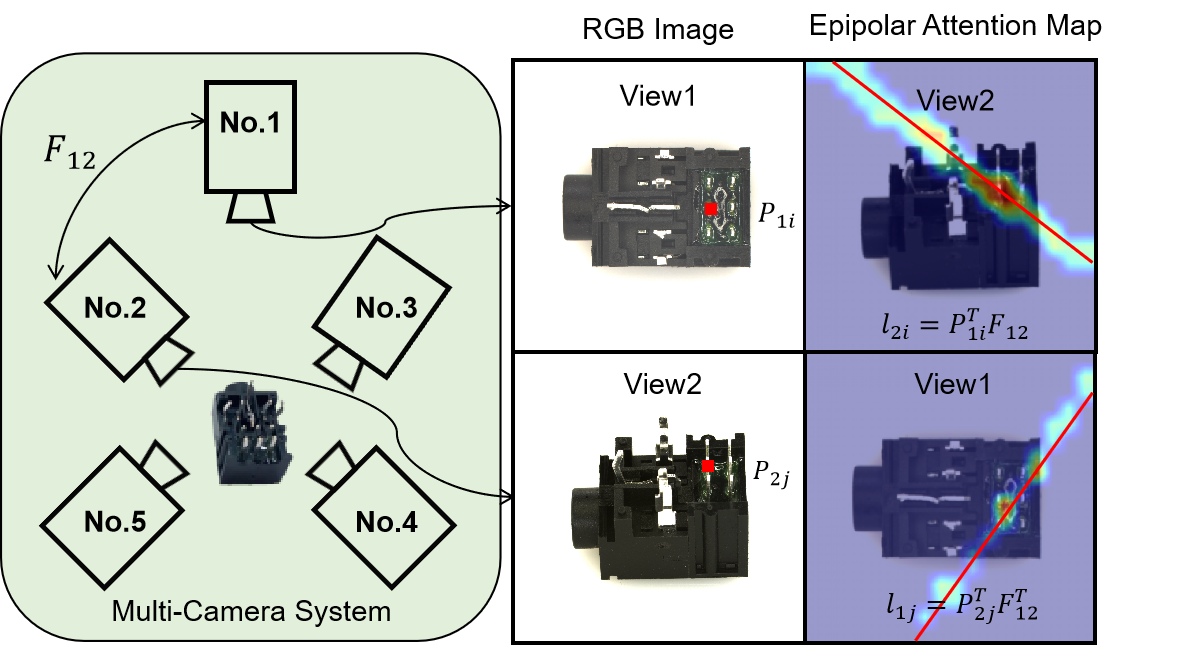}
    \label{fig:short-a}
  \vspace{-0.4cm}
  \caption{An illustration of epipolar geometry guided attention in a multi-camera system. The corresponding patch of $P_{1i}$ in views lies on the epipolar line $l_{2i}=P^\top_{1i}F_{12}$~(red line). Attention map is restricted to the patches on the epipolar line.}
  \label{fig:epi_vis}
  \vspace{-0.4cm}
\end{figure}

Despite the advantages of multi-view observations, existing approaches often fail to fully exploit cross-view relationships~\cite{wang2024real}. A common strategy is to train separate models for each viewpoint and perform inference independently, neglecting the correlation between views. This limits the potential of multi-view information, as defects may be subtle in certain perspectives while more pronounced in others. Effectively fusing information across viewpoints could improve anomaly detection by leveraging complementary features, thus improving anomaly prediction at the level of each view respectively and the overall object.

Recent efforts in multi-view anomaly detection have explored feature fusion, cross-view feature alignment, and self-supervised learning~\cite{wang2022multiview}. However, many of these methods treat multi-view data similarly to multi-modal inputs, which may not be optimal for structured multi-view camera systems. Inspired by the success of attention-based fusion~\cite{wang2018non}, the Multi-View Adaptive Selection (MVAS) attention mechanism was proposed to enhance multi-view feature aggregation in MVIAD~\cite{he2024learning}. MVAS partitions feature maps into patches and employs cross-attention to aggregate features from corresponding locations across views. The selection of relevant patches relies on semantic similarity, with cross-view attention learned through reconstruction losses under a reverse distillation framework~\cite{deng2022anomaly}. The success of MVAS highlights the importance of accurately identifying corresponding regions across views for effective feature fusion.

\begin{figure*}[!htb]
    \centering
    \includegraphics[width=\linewidth]{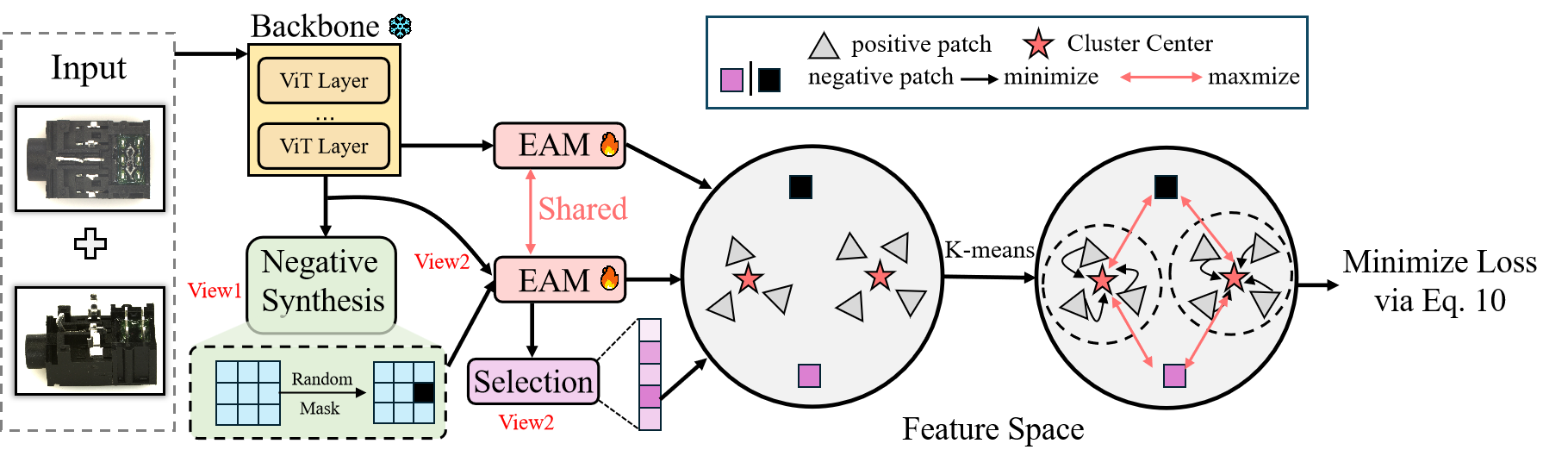}
    \label{fig:short-a}
    \vspace{-0.6cm}
  \caption{
  Overview of the proposed multi-center pre-training framework. Multi-view images are processed by a frozen backbone, followed by cross-view fusion via the EAM module to construct a patch-level feature bank. Features from reference view are randomly masked, while other views remain unchanged. In the EAM output, masked patches from the reference view and patches explicitly selected from source views via \cref{eq:neg_samp} serve as negatives, with the loss computed using \cref{eq:final_loss}.
  }.
  \vspace{-0.6cm}
  \label{fig:pipeline}
\end{figure*}

Despite these advances, existing methods overlook a fundamental geometric constraint in multi-view camera systems. As illustrated in ~\cref{fig:epi_vis}, industrial inspection setups often have significant overlap between camera views, meaning observations are constrained by epipolar geometry~\cite{hartley2003multiple}. Specifically, for a pixel \( p_{1i} \) in view 1, its corresponding projection in view 2, \( p_{2i} \), must satisfy the epipolar constraint,  \(p_{1i}^\top F_{12} p_{2i} = 0
\), where \( F_{12} \) is the fundamental matrix. This suggests that cross-view attention should not rely solely on semantic similarity~\cite{he2024learning} but could be guided by epipolar geometry to enhance feature correspondence. To this end, we propose an \textbf{Epipolar Attention Module (EAM)} for cross-view feature fusion. Given a reference patch in one view, we first identify candidate corresponding patches in other views along the epipolar line. These patches then serve as keys and values in cross-view attention, ensuring feature aggregation is geometrically consistent. We append EAM to the output of regular ViT encoder to fuse features across multiple viewpoints. Our anomaly detection framework is built upon a memory bank based method~\cite{roth2022towards} due to the compatibility with arbitrary encoder architectures, in comparison, distillation-based methods~\cite{he2024learning,deng2022anomaly,tien2023revisiting} are mostly restricted to CNN architectures. Therefore, our proposed method is referred to as \textbf{M}ulti-\textbf{V}iew \textbf{E}pipolar \textbf{A}ttention Anomaly \textbf{D}etection~(\textbf{MVEAD}) with an overview presented in Fig.~\ref{fig:pipeline}.

Our experiments in Tab.~\ref{tab:main_ablation} reveal that EAM is ineffective when the projection layers are not properly initialized. Existing attempts to integrate epipolar geometry into cross-view attention optimizes the projection layers in a fully supervised manner~\cite{he2020epipolar}. However, memory-bank-based anomaly detection lacks an explicit objective for training attention projection layers. To address this challenge, we introduce a \textit{multi-center pretraining strategy}, inspired by one-class classification which minimizes feature space volume~\cite{ruff2018deep}. Given the complex distribution of normal patterns in industrial images, we extend this idea by introducing multiple cluster centers and reduce the gap between training samples and its closest cluster center. To prevent model collapse, where all features map to a single or few centers, we introduce \textit{synthetic negative samples} tailored to epipolar attention properties as a regularization mechanism. The negative sample synthesis differs from the purely synthesis-based anomaly detection methods~\cite{li2021cutpaste,zhang2024realnet} in that we focus more on the interaction between different viewpoints for synthesis and use them as regularization. Finally, due to view-dependent defect variations, a single memory bank is suboptimal for multi-view anomaly detection. To improve inference, we use a multi-view memory bank strategy to maintain separate banks for each viewpoint.


We summarize the key contributions of this work as follows:  

\begin{itemize}
    \item We recognize that defects exhibit varying visibility across different viewpoints in a multi-view camera system. While existing methods rely solely on data-driven feature fusion, we propose leveraging epipolar geometry to guide cross-view attention, ensuring more precise feature aggregation.  
    \item We extend the memory bank-based anomaly detection framework, which lacks an explicit training objective, by introducing a multi-center pretraining strategy to optimize the parameters of the epipolar attention module. Additionally, we incorporate consistency-based negative sample synthesis to prevent trivial solutions.  
    \item We validate our approach on the state-of-the-art multi-view industrial anomaly detection dataset, demonstrating its effectiveness while maintaining inference efficiency.  
\end{itemize}



\section{Related Work}

\noindent\textbf{Industrial Anomaly Detection}: Anomaly detection (AD) \cite{cao2024survey} identifies anomalies in target-class images and localizes anomaly regions. Mainstream AD techniques fall into three categories. Synthesizing-Based Approaches introduce artificial defects to expose models to diverse anomalies. CutPaste \cite{li2021cutpaste} relocates patches within an image, while SimpleNet \cite{liu2023simplenet} perturbs feature space with Gaussian noise. DRAEM \cite{zavrtanik2021draem} synthesizes anomalies using texture datasets \cite{cimpoi2014describing}, and RealNet \cite{zhang2024realnet} improves anomaly realism with diffusion models \cite{ho2020denoising}.  
Reconstruction-Based Approaches assume models trained on normal data struggle to reconstruct anomalies. AutoEncoders \cite{akcay2019ganomaly}, GANs \cite{schlegl2019f}, VAEs \cite{an2015variational}, and diffusion models \cite{zhang2024realnet} are commonly used. However, these models may fail if anomalies resemble nominal data or exploit shortcut learning \cite{zavrtanik2021reconstruction}. Recent methods \cite{zavrtanik2021draem,collin2021improved, liao2024coft} address this by injecting pseudo anomalies or using additional supervision.  Embedding-Based Approaches learn feature representations where nominal data cluster compactly, and anomalies deviate. PaDiM \cite{defard2021padim} models local feature statistics, PatchCore \cite{roth2022towards} detects anomalies via nearest-neighbor search in a memory bank, and knowledge distillation methods \cite{deng2022anomaly} compare outputs from teacher-student networks. Normalizing Flow \cite{kobyzev2020normalizing} \cite{rudolph2021same} models feature distributions but incurs high computational costs due to limited downsampling.

\noindent\textbf{Multi-View Deep Learning}:
Multi-view representation learning has been widely applied across various domains. Early work such as MVCNN \cite{su2015multi} pioneered CNN-based multi-view aggregation. In pose estimation, multi-view fusion enhances feature representation \cite{qiu2019cross,ma2022ppt}, with AdaFuse \cite{zhang2021adafuse} selecting the highest heatmap response along the epipolar line and the Epipolar Transformer \cite{he2020epipolar} leveraging a non-local module \cite{wang2018non} for adaptive fusion.
For multi-view stereo (MVS) \cite{furukawa2015multi}, MVSNet \cite{gu2020cascade} introduced cost volume regularization for scene reconstruction, inspiring further refinements in efficiency and depth accuracy \cite{sheng2019multi,xu2020pvsnet,wang2021patchmatchnet}.

\noindent\textbf{Multi-View Anomaly Detection}: Multi-view anomaly detection remains underexplored. Existing works addressing multi-view anomaly detection have studied multiple modality inputs~\cite{wang2020towards,wang2022multiview}, different lightining conditions~\cite{liu2024learning}, etc. Industrial anomaly detection from multiple camera viewpoints was less studied due to the lack of valid benchmarks. Real-IAD \cite{wang2024real} introduced an initial dataset featuring top-down and side view observations of industrial objects with extensive evaluation of state-of-the-art single-view industrial anomaly detection methods. To exploit complementary information from multiple views, MVAD~\cite{he2024learning} applied neighborhood attention for feature fusion across views. The attention is computed via semantic similarity, thus lacking explicit geometric constraints. Our method improves upon this by incorporating an epipolar transformer-inspired attention mechanism, leveraging epipolar geometry for more precise and robust multi-view feature aggregation.


\section{Methods}

\subsection{Preliminaries}

We formally define the task of multi-view anomaly detection as follows. Each instance consists of images captured from \( V \) viewpoints, denoted as \( x_i \in \mathbb{R}^{V \times 3 \times H \times W} \). The objective is to predict anomalies at both the instance and pixel levels, identifying defective regions in each view.  

\noindent\textbf{Memory Bank-Based Anomaly Detection}:
We revisit the memory bank-based anomaly detection paradigm. Given a pre-trained encoder network \( f(x; \Theta) \), such as a ViT \cite{dosovitskiy2020image}, the input image is first patchified into tokens, which are then processed through transformer blocks to extract tokenized features. For an input sample \( x_i \), we denote its tokenized feature on the $v$-th view as \( z_{iv} \in \mathbb{R}^{H \cdot W / P^2 \times D} \), where \( D \), \( H \), \( W \), and \( P \) represent the feature channels, input image height, width, and patch size, respectively.  

A memory bank \( \mathcal{M} = \{m_k\} \) stores representative tokenized features for anomaly prediction. Since storing all tokens from normal training data is impractical due to prohibitive memory requirements, core-set selection is typically employed to retain a subset of tokens in \( \mathcal{M} \)~\cite{roth2022towards}. Anomalies in a test image are detected at the token level using the following scoring function:  

\begin{equation}\label{eq:pc_inf}
    s_{i}=\max_j\min_{m_k\in \mathcal{M}} ||z_{ivj} - m_k||
\end{equation}  

\noindent\textbf{Cross-View Attention for Multi-View Anomaly Detection}:
Detecting anomalies using multi-view observations is an emerging topic. Early approaches formulated this task by independently predicting anomalies for each view~\cite{wang2024real}, failing to leverage the strong correlations between views. To exploit cross-view information, \cite{he2024learning} introduced a cross-view attention mechanism, aggregating information from supporting views for each reference view.  

Specifically, we refer \( z_i \) to as a tokenized feature map \( z_i \in \mathbb{R}^{V \times H \cdot W / P^2 \times D} \) extracted by a transformer network. The attention mechanism between the reference view \( z_{ia} \) and supporting views \( \{z_{ib}\}_{b\neq a} \) follows a standard cross-attention formulation, where \( W_Q \), \( W_K \), and \( W_V \) are learnable projection matrices,

\vspace{-0.3cm}
\begin{equation}
\begin{split}
    &Attn(z_{ia})=\text{softmax}\left(\frac{Q K^\top}{\sqrt{D}}\right),\\
    &Q=W_Qz_{ia}, \quad K=W_Kz_{ib}, \quad V=W_Vz_{ib}
\end{split}
\end{equation}  

While cross-view attention improves upon independent predictions, it relies solely on semantic similarity, which can be unreliable for textureless or repetitive patterns. Learning such attention in a purely data-driven manner demands large-scale training data and is prone to errors. To address this, we propose leveraging epipolar geometry as a structural constraint, enhancing both robustness and efficiency in cross-view fusion.



\subsection{Epipolar Attention for Multi-View Fusion}  
\label{sec:epiattn}  

\noindent\textbf{Epipolar Geometry}:  
In non-degenerate stereo vision, pixel correspondences between two camera views are governed by epipolar geometry. Given a pixel in view \( a \), denoted as \( p_a = [u_a, v_a, 1]^\top \), its corresponding projection in view \( b \), \( p_b = [u_b, v_b, 1]^\top \), must lie on the epipolar line, satisfying the epipolar constraint:  

\begin{equation}
    p_a^\top F_{ab} p_b = 0
\end{equation}  

where \( F_{ab} \) is the fundamental matrix relating the two views. The corresponding epipolar line in view \( b \) is parameterized as \( l_b = F_{ab}^\top p_a \). The fundamental matrix \( F_{ab} \) can be efficiently estimated using the eight-point algorithm~\cite{hartley2003multiple}.

\noindent\textbf{Cross-View Attention with Epipolar Constraint}:
Unlike existing cross-view attention mechanisms that rely solely on feature similarity, we propose incorporating epipolar geometry as a structural prior to guide cross-view attention. Specifically, given a patch in the reference view \( z_{iaj} \), we first determine its center point \( p_{iaj} \) using:

\begin{equation}\label{eq:center_point}
p_{iaj}=
\begin{bmatrix} 
\left( \left \lfloor \frac{jP}{W} \right \rfloor + 0.5 \right) P \\ 
\left( j \bmod \frac{W}{P} + 0.5 \right) P \\ 
1
\end{bmatrix}
\end{equation}

Following epipolar geometry, the projection of \( p_{iaj} \) in the \( b \)-th supporting view lies on the epipolar line parameterized by \( (p_{iaj}^\top F_{ab}) p_{ibk} = 0 \). Any \( p_{ibk} \) satisfying this equation could correspond to \( p_{iaj} \). To integrate this prior into cross-view attention, we identify all patches in the supporting view that overlap with the epipolar line. We define a binary mask \( M_{ab} \in \{0,1\}^{H\cdot W/P^2\times H\cdot W/P^2} \) that dictates the epipolar attention between views \( a \) and \( b \), formulated as,

\vspace{-0.2cm}
\begin{equation}\label{eq:epattn_mask}
\begin{split}
M_{ab}[j,k] &= \begin{cases} 
1, & \text{if } d_{jk} \leq \delta \\ 
0, & \text{otherwise} 
\end{cases} \\
\text{where} \quad d_{jk} &= \frac{| l_b p_{ibk} |}{\sqrt{(l_b)_1^2 + (l_b)_2^2}}, \quad l_b = p_{iaj}^\top F_{ab}
\end{split}
\end{equation}

The final Epipolar Attention Module (EAM) is formulated as,
\begin{equation}\label{eq:epattn}
\begin{split}
    &z_{iaj} = W_O \sum_{k} M_{ab}[j,k] \cdot Attn(z_{iaj}) V + z_{iaj}, \\
    &Attn(z_{iaj}) = \frac{\exp(QK^\top/\sqrt{D})}{\sum_{k} \exp(QK_k^\top/\sqrt{D}) M_{ab}[j,k]}, \\
    &Q = W_Q z_{iaj}, \quad K = W_K z_{ib}, \quad V = W_V z_{ib}.
\end{split}
\end{equation}

During pre-training and inference, the EAM module takes the output of the \( k \)-th layer of the ViT-like backbone as input, generating the feature for constructing the memory bank while facilitating feature aggregation across all supporting views and leveraging epipolar constraints for improved robustness and efficiency.

\begin{figure}[!htb]
  \vspace{-0.3cm}
  \centering
    \centering
    \includegraphics[width=\linewidth]{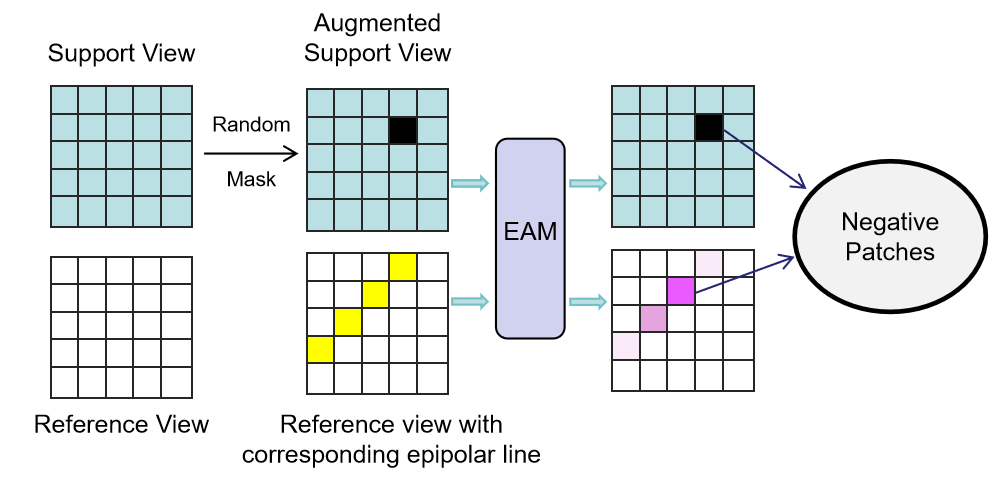}
    \vspace{-0.6cm}
  \caption{After randomly selecting the support view, a random mask is applied. Following EAM processing, the Top-K patches along the epipolar line corresponding to the masked region in the reference view are selected based on \cref{eq:neg_samp} and used as negative patches along with the masked patch.}
  \vspace{-0.4cm}
  \label{fig:neg_vis}
\end{figure}


\subsection{Multi-Center Pre-training}
\label{sec:method_pretrain}  

Unlike epipolar transformer~\cite{he2020epipolar}, where end-to-end training is available to refine linear projection layers, memory bank-based anomaly detection lacks an explicit training objective for optimizing the backbone network. Leaving these projection layers randomly initialized can degrade performance. To address this, we propose a pre-training strategy for optimizing the projection layers in EAM.  

Inspired by unsupervised one-class classification, DeepSVDD~\cite{ruff2018deep} pre-trains networks by minimizing the volume of a data-enclosing hypersphere with radius \( R>0 \) and center \( c\in\mathbb{R}^D \). However, a single feature cluster may be insufficient for modeling industrial objects with diverse appearances. Since the inference relies on nearest-neighbor search in the memory bank, a well-distributed set of prototypes should follow a multi-modal distribution. This motivates our multi-center pre-training approach, which preserves multiple feature centers.  
We maintain \( K \) cluster centers \( \{c_k\in\mathbb{R}^D\}_{k=1\cdots K} \) in the patch feature space, initializing them via K-Means clustering. During pre-training, patches are iteratively assigned to their closest prototype, and their distances are minimized:  

\vspace{-0.3cm}
\begin{equation}\label{eq:pos_loss}
\begin{split}
    &\min_\Theta \frac{1}{N}\sum_{i,v,j} ||z_{ivj} - m_{k^*}||,\\
    s.t.\quad& {k^*}=\arg\min_{m_{k}\in\mathcal{M}} ||m_{k} - z_{ivj}||
\end{split}
\end{equation}  

\noindent\textbf{Multi-View Memory Bank}
Existing memory bank-based methods primarily target single-view datasets, maintaining a single memory bank for normal sample distributions. In multi-view anomaly detection, significant viewpoint changes can obscure defect discrimination. To address this, we propose a multi-view memory bank, $\{\set{M}_v\}_{v=1\cdots V}$, where each individual view maintains its own memory representation.

\noindent\textbf{Regularization with Synthesized Negative Samples}: 
Pre-training solely on clean patches risks model collapse, compressing all features into a few trivial representations. To mitigate this, we introduce negative sample regularization, synthesizing anomalous patches to enforce discriminative learning.  

Unlike single-view setups, multi-view anomaly detection requires leveraging cross-view consistency to generate meaningful negative samples. For each sample with \( V \) viewpoints, we randomly select one viewpoint as the support view b. 
Given a reference view \( a \) and the support view \( b \), we follow the steps to identify negative patches.

First, we select a patch \( p_{ibk} \) in view \( b \) and identify corresponding patches in view \( a \) using epipolar constraints (\cref{sec:epiattn}).  
We further extract reference view features \( Z_{ia}=\{z_{iaj}\} \) and then mask out the support view patch and perform inferene again, obtaining updated features \( \tilde{Z}_{ia}=\{\tilde{z}_{iaj}\} \).  
Finally, we identify potential negative patches \( \set{Z}_{neg} \) by selecting the top \( N_k \) most altered patches,
\vspace{-0.3cm}
\begin{equation}\label{eq:neg_samp}
\vspace{-0.1cm}
\begin{split}
    \tilde{\set{Z}}_{neg}&=\{\tilde{z}_{iaj}\}_{j=1\cdots J^*[1:N_k]}\cup \left \{ \tilde{z}_{ibk}\right \} ,\\
    s.t.\quad J^*&=\arg\text{sortd}_j||z_{iaj}-\tilde{z}_{iaj}||_2
\end{split}
\end{equation}  

Negative patches are incorporated into pre-training by penalizing their logarithmic average distance to memory bank cluster centers, preventing them from collapsing into trivial solutions,

\vspace{-0.3cm}
\begin{equation}
\vspace{-0.1cm}
    \mathcal{L}_{neg}(z_{iv})=-\sum_{\tilde{z}_{ivj}\in\tilde{\set{Z}}_{neg}} \log\left(\frac{1}{|\set{M}_v|}\sum_{m_k\in\set{M}_v} ||\tilde{z}_{ivj}-m_k||_2\right)
\end{equation}  

\noindent\textbf{Final Training Objective}:
Our final loss combines positive and negative patches,

\vspace{-0.3cm}
\begin{equation}\label{eq:final_loss}
\vspace{-0.1cm}
\begin{split}
 &\min_\Theta \sum_{i,v,j} ||z_{ivj} - m_{k^*}|| + \lambda\sum_{i,v}\mathcal{L}_{neg}(z_{i}),\\
    &s.t.\quad {k^*}=\arg\min_{m_{k}\in\mathcal{M}_v} ||m_{k} - z_{ivj}||
\end{split}
\end{equation}  


\noindent\textbf{Inference}:
The sample-level anomaly score is determined as the maximum across all views, with each view’s anomaly score being the highest patch-level deviation from the nearest memory bank prototype,

\vspace{-0.2cm}
\begin{equation}
\vspace{-0.1cm}
s_{i}=\underset{0\le i< v}{\max} \ s_{iv},\quad     s_{iv}=\max_j\min_{m_k\in \mathcal{M}_v} ||z_{ivj} - m_k||
\label{eq:inference}
\end{equation}

\section{Experiments}

\subsection{Experiment Details}

\begin{figure*}[!hbt]
  \centering
  \begin{subfigure}{\linewidth}
    \centering
    \includegraphics[width=\linewidth]{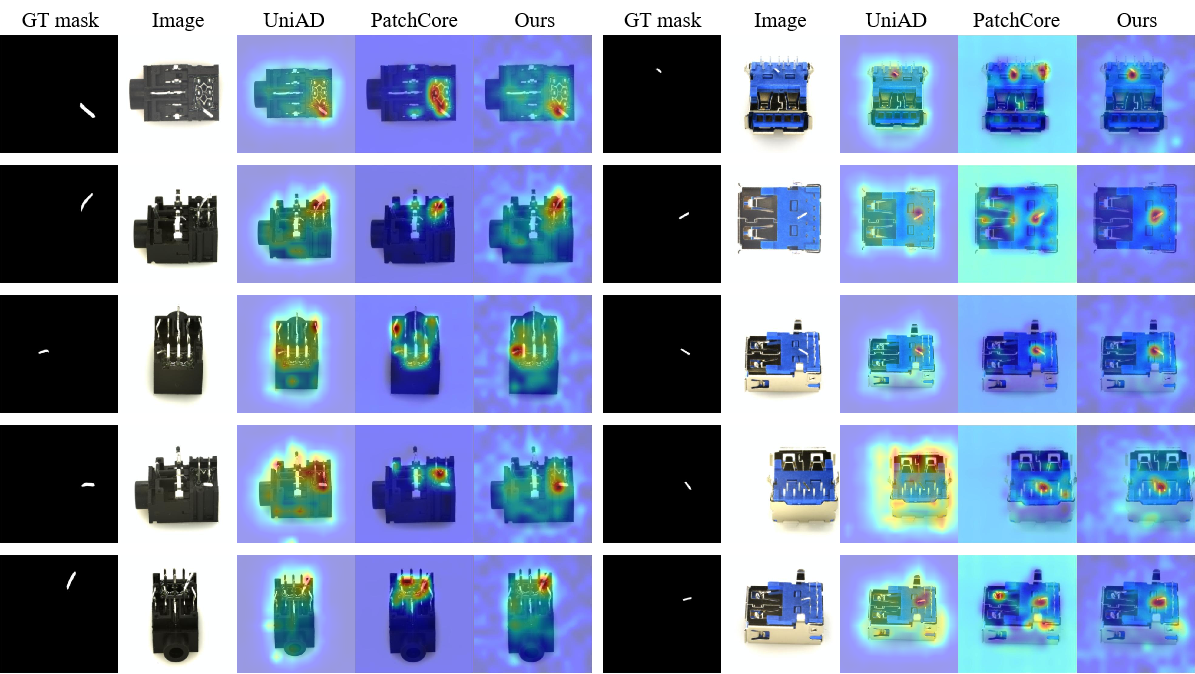}
    \label{fig:short-a}
  \end{subfigure}
  \vspace{-0.8cm}
  \caption{Anomaly segmentation results for two samples from the Real-IAD dataset: the audiojack class (left) and the USB class (right).}
  \vspace{-0.6cm}
  \label{fig:vis_sample}
\end{figure*}

\noindent\textbf{Datasets}:
To evaluate multi-view industrial anomaly detection, we conduct experiments on Real-IAD~\cite{wang2024real}, a dataset comprising 30 sub-datasets with a total of 150K high-resolution images, each representing a distinct industrial object. For each object, images are captured from five different viewpoints to simulate real-world inspection conditions. Following standard unsupervised anomaly detection protocols, we use only normal samples for training, while test sets contain both normal and anomalous images with pixel-wise anomaly annotations.
Our evaluation includes both single-class and multi-class detection settings. i) Single-class setting: Models are trained and evaluated independently on each object category, ensuring a fair per-category assessment. ii) Multi-class setting: The full dataset is used for pre-training and memory bank construction, enabling evaluation in a broader setting.

\begin{table*}[t]
  \centering
  \caption{Comparison of \textbf{Sample-level AUROC} and \textbf{Sample-level AP} results for our method on Real-IAD dataset with \textbf{Single-class/Multi-class setting}.}
  \vspace{-0.3cm}
  \setlength{\tabcolsep}{3pt}
  \resizebox{1\linewidth}{!}{
    \begin{tabular}{lccccc|ccccc}
      \toprule
      \multicolumn{1}{c}{} & \multicolumn{5}{c}{AUROC} & \multicolumn{5}{c}{AP } \\
      \cmidrule(lr){2-6} \cmidrule(lr){7-11}
      Category & UniAD\cite{you2022unified} & SimpleNet\cite{liu2023simplenet} & MVAD\cite{he2024learning} & PatchCore\cite{roth2022towards} & MVEAD~(Ours) & UniAD\cite{you2022unified} & SimpleNet\cite{liu2023simplenet} & MVAD\cite{he2024learning} & PatchCore\cite{roth2022towards} & MVEAD~(Ours) \\
            \midrule
        audiojack & 88.4/90.3 & 93.0/68.3 & 93.1/91.2 & 93.0/\textbf{93.2} & \textbf{95.3}/92.7 & 93.4/95.2 & 96.5/82.4 & 96.3/94.9 & 96.5/\textbf{96.6} & \textbf{97.0}/96.4 \\
        bottle cap & 90.8/97.2 & 99.3/51.2 & 98.9/\textbf{97.8} & 95.1/93.9 & \textbf{99.5}/93.5 & 95.2/98.7 & \textbf{99.7}/70.4 & 99.5/\textbf{98.9} & 97.8/97.3 & \textbf{99.7}/97.1 \\
        button battery & 88.3/95.5 & 96.1/57.2 & 93.4/80.0 & 93.3/94.2 & \textbf{97.0}/\textbf{97.5} & 91.8/95.1 & 96.4/75.2 & 96.4/90.4 & 95.5/96.8 & \textbf{96.8}/\textbf{97.2} \\
        end cap & 85.3/87.7 & 95.1/55.7 & 91.9/86.6 & 94.5/90.6 & \textbf{95.6}/\textbf{92.0} & 92.0/94.2 & 97.7/73.7 & 96.1/93.0 & 97.3/95.3 & \textbf{97.8}/\textbf{95.9} \\
        eraser & 93.2/90.1 & 94.3/35.8 & 91.8/87.2 & 94.3/91.5 & \textbf{95.8}/\textbf{92.5} & 96.8/95.3 & \textbf{97.5}/61.6 & 96.3/94.3 & 97.4/96.1 & 97.2/\textbf{96.6} \\
        fire hood & 89.4/85.9 & 93.5/54.2 & 90.2/81.8 & 85.1/82.3 & \textbf{93.8}/\textbf{91.8} & 94.1/92.9 & 96.5/69.4 & 95.1/90.7 & 92.5/90.6 & \textbf{97.3}/\textbf{95.4} \\
        mint & 61.0/66.5 & 86.1/53.1 & 85.9/68.1 & 84.6/82.3 & \textbf{87.4}/\textbf{86.5} & 88.0/89.7 & 96.7/84.7 & 96.5/91.0 & 96.3/95.2 & \textbf{96.9}/\textbf{96.7} \\
        mounts & 93.4/98.3 & 99.5/64.1 & \textbf{99.6}/98.8 & 98.9/98.6 & 99.1/\textbf{98.9} & 96.9/99.2 & 99.7/79.7 & \textbf{99.8}/\textbf{99.5} & 99.5/99.4 & 99.5/99.4 \\
        pcb & 79.4/83.8 & 91.5/61.2 & 91.1/\textbf{90.0} & 91.1/89.2 & \textbf{92.4}/88.8 & 86.7/91.6 & 95.7/77.2 & 95.9/\textbf{95.3} & 95.9/95.1 & \textbf{96.4}/95.0 \\
        phone battery & 91.5/85.4 & 96.3/62.4 & 94.2/92.7 & 94.7/94.3 & \textbf{96.6}/\textbf{96.1} & 96.1/93.1 & 98.1/78.4 & 97.2/96.7 & 97.4/97.2 & \textbf{98.3}/\textbf{98.1} \\
        plastic nut & 86.4/84.7 & 96.1/48.2 & \textbf{97.2}/90.8 & 95.6/94.2 & 96.4/\textbf{95.2} & 88.8/89.5 & 97.5/66.3 & \textbf{98.5}/94.7 & 97.4/\textbf{97.1} & 97.9/97.0 \\
        plastic plug & 65.6/80.2 & \textbf{96.0}/50.2 & 94.4/89.1 & 94.5/92.3 & 95.4/\textbf{92.8} & 81.2/90.3 & \textbf{98.2}/71.0 & 97.3/94.6 & 97.6/96.6 & 98.0/\textbf{96.8} \\
        porcelain doll & 65.9/90.1 & 96.6/80.2 & 96.2/94.9 & 97.5/96.2 & \textbf{98.5}/\textbf{97.1} & 78.4/91.9 & 98.2/89.1 & 98.3/97.3 & 98.5/98.2 & \textbf{99.4}/\textbf{98.6} \\
        regulator & 52.1/65.7 & 96.3/49.9 & 87.7/73.6 & \textbf{97.2}/85.0 & 96.7/\textbf{91.5} & 69.8/80.6 & 98.5/68.1 & 93.5/86.8 & \textbf{98.6}/92.2 & 98.4/\textbf{95.2} \\
        rolled strip base & 97.8/98.6 & \textbf{99.8}/65.5 & 99.6/97.7 & 99.2/97.8 & 99.7/\textbf{99.2} & 99.0/99.3 & \textbf{99.9}/80.8 & 99.8/98.7 & 99.6/98.8 & 99.5/\textbf{99.6} \\
        sim card set & 93.2/87.3 & 99.0/77.1 & 98.2/96.2 & 98.6/97.5 & \textbf{99.5}/\textbf{99.3} & 93.6/90.8 & \textbf{99.5}/86.7 & 99.1/97.9 & 99.4/98.7 & \textbf{99.5}/\textbf{99.7} \\
        switch & 92.9/91.5 & 98.6/66.8 & 96.3/94.9 & 98.6/\textbf{97.8} & \textbf{98.7}/96.7 & 96.4/96.2 & \textbf{99.4}/82.0 & 98.4/97.6 & \textbf{99.4}/\textbf{98.9} & \textbf{99.4}/98.5 \\
        tape & 94.8/98.2 & 99.9/54.2 & \textbf{100.0}/98.4 & 99.9/99.2 & 99.8/\textbf{99.5} & 97.4/99.0 & \textbf{100.0}/73.9 & \textbf{100.0}/99.3 & \textbf{100.0}/99.6 & 99.9/\textbf{99.7} \\
        terminalblock & 82.9/95.7 & 98.2/75.4 & 98.6/96.8 & \textbf{98.7}/97.9 & 98.5/\textbf{98.4} & 92.2/98.2 & 99.3/87.8 & 99.4/98.7 & \textbf{99.5}/99.2 & 99.4/\textbf{99.3} \\
        toothbrush & 94.3/89.8 & 96.1/71.2 & 95.8/87.0 & \textbf{97.7}/96.0 & 97.0/\textbf{96.5} & 96.6/92.9 & 98.1/83.7 & 97.0/93.1 & \textbf{98.5}/\textbf{97.7} & 98.3/97.5 \\
        toy & 86.3/75.4 & 93.8/59.0 & 93.6/86.4 & 97.0/93.1 & \textbf{97.4}/\textbf{97.5} & 90.9/85.7 & 96.9/74.3 & 96.8/92.3 & \textbf{98.6}/95.8 & \textbf{98.6}/\textbf{98.8} \\
        toy brick & 80.1/78.6 & 87.1/59.2 & 80.8/69.4 & 81.3/76.1 & \textbf{88.3}/\textbf{84.8} & 85.6/84.3 & 92.6/73.6 & 89.7/81.7 & 90.0/86.1 & \textbf{93.6}/\textbf{91.7} \\
        transistor1 & 96.1/98.4 & \textbf{99.8}/66.8 & \textbf{99.8}/\textbf{98.8} & 98.7/98.2 & \textbf{99.8}/98.4 & 97.7/99.2 & \textbf{99.9}/81.6 & \textbf{99.9}/\textbf{99.5} & 99.3/99.1 & 99.5/99.2 \\
        u block & 91.2/94.8 & \textbf{98.7}/50.5 & 98.4/93.0 & 98.0/\textbf{98.2} & 98.1/97.3 & 94.9/97.0 & \textbf{99.4}/75.3 & 99.0/96.1 & 98.9/\textbf{99.1} & 98.4/98.3 \\
        usb & 84.0/79.7 & 93.9/64.1 & \textbf{96.3}/92.1 & 95.1/\textbf{94.1} & 95.2/92.7 & 88.9/88.4 & 96.1/79.9 & \textbf{97.7}/95.9 & 96.8/\textbf{96.7} & 96.3/96.0 \\
        usb adaptor & 78.4/82.9 & 93.8/48.8 & 93.3/91.0 & 94.9/92.6 & \textbf{95.4}/\textbf{93.8} & 87.7/90.3 & 97.1/70.4 & 96.5/95.5 & 97.8/96.7 & \textbf{98.1}/\textbf{97.2} \\
        vcpill & 79.8/80.7 & 96.8/64.5 & 95.9/88.2 & \textbf{98.2}/96.2 & 97.8/\textbf{97.1} & 88.9/89.9 & 98.4/78.7 & 98.1/94.2 & \textbf{99.2}/98.2 & 98.9/\textbf{98.5} \\
        wooden beads & 78.5/77.3 & 92.1/59.5 & 92.7/82.2 & 93.9/90.8 & \textbf{95.7}/\textbf{94.4} & 90.1/90.1 & 97.0/81.3 & 97.2/92.7 & 97.7/96.2 & \textbf{98.4}/\textbf{98.0} \\
        woodstick & 81.8/84.0 & 80.3/58.2 & 85.0/76.8 & 84.2/79.4 & \textbf{86.0}/\textbf{84.7} & 87.3/90.7 & 90.5/72.7 & 92.5/87.5 & 92.2/89.6 & \textbf{92.9}/\textbf{92.7} \\
        zipper & 94.6/97.8 & 99.8/91.9 & 99.6/\textbf{99.9} & 99.8/99.8 & \textbf{100.0}/\textbf{99.9} & 96.2/98.6 & \textbf{99.9}/96.0 & 99.8/\textbf{99.9} & \textbf{99.9}/\textbf{99.9} & \textbf{99.9}/\textbf{99.9} \\
        \midrule
        Average & 84.6/87.1 & 94.6/60.8 & 94.3/89.0 & 94.8/92.8 & \textbf{96.2}/\textbf{94.6} & 91.1/92.9 & 97.3/77.5 & 97.3/94.6 & 97.5/96.5 & \textbf{98.0}/\textbf{97.3} \\
        
      \bottomrule
    \end{tabular}
  }
  \label{tab:sample_results}
\end{table*}

\begin{table*}[t]
  \centering
  \caption{Comparison of \textbf{Image-level AUROC} and \textbf{Image-level AP} results for our method on Real-IAD dataset with \textbf{Single-class/Multi-class setting}.}
    \vspace{-0.3cm}
  \setlength{\tabcolsep}{3pt}
  \resizebox{1\linewidth}{!}{
    \begin{tabular}{lccccc|ccccc}
      \toprule
      \multicolumn{1}{c}{} & \multicolumn{5}{c}{AUROC} & \multicolumn{5}{c}{AP } \\
      \cmidrule(lr){2-6} \cmidrule(lr){7-11}
      Category & UniAD\cite{you2022unified} & SimpleNet\cite{liu2023simplenet} & MVAD\cite{he2024learning} & PatchCore\cite{roth2022towards} & Ours & UniAD\cite{you2022unified} & SimpleNet\cite{liu2023simplenet} & MVAD\cite{he2024learning} & PatchCore\cite{roth2022towards} & Ours \\
            \midrule
              
        audiojack & 78.7/81.4 & 88.4/58.4 & 86.9/80.3 & 86.5/86.3 & \textbf{90.1}/\textbf{87.7} & 60.8/76.6 & 84.4/44.2 & 82.5/72.8 & 82.8/82.7 & \textbf{87.5}/\textbf{84.1}\\
        bottle cap & 85.6/\textbf{92.5} & 91.1/54.1 & \textbf{95.6}/92.4 & 93.7/88.2 & 95.1/87.8 & 82.6/\textbf{91.7} & 88.6/47.6 & \textbf{95.4}/90.9 & 93.3/89.5 & 94.2/88.1\\
        button battery & 65.9/75.9 & 88.4/52.5 & 90.8/76.6 & 86.8/88.3 & \textbf{91.2}/\textbf{90.9} & 71.9/81.6 & 89.9/60.5 & 92.1/83.2 & 89.5/\textbf{90.7} & \textbf{94.4}/\textbf{90.7}\\
        end cap & 80.6/80.9 & 83.7/51.6 & 85.8/79.4 & 84.8/81.0 & \textbf{86.9}/\textbf{81.3} & 84.4/86.1 & 88.4/60.8 & 88.8/84.6 & 88.1/86.1 & \textbf{88.9}/\textbf{86.2}\\
        eraser & 87.9/90.3 & 91.6/46.4 & 91.2/88.6 & 92.5/90.0 & \textbf{92.9}/\textbf{90.6} & 82.4/\textbf{89.2} & 90.1/39.1 & 89.2/87.2 & \textbf{90.4}/88.7 & \textbf{90.4}/88.6\\
        fire hood & 79.0/80.6 & 81.7/58.1 & 84.6/78.6 & 80.1/80.3 & \textbf{87.5}/\textbf{87.0} & 72.3/74.8 & 74.1/41.9 & 77.2/71.8 & 75.3/73.6 & \textbf{83.6}/\textbf{80.1}\\
        mint & 64.5/67.0 & 77.0/52.4 & 79.5/68.9 & 76.5/75.3 & \textbf{82.3}/\textbf{76.4} & 63.8/66.6 & 78.5/50.3 & 80.7/70.2 & 78.1/75.4 & \textbf{82.0}/\textbf{76.9}\\
        mounts & 84.1/87.6 & \textbf{88.2}/58.7 & 87.9/\textbf{89.5} & 86.8/87.2 & 88.0/84.8 & 71.2/77.3 & \textbf{79.1}/48.1 & 75.6/\textbf{81.5} & 74.6/77.8 & 76.3/72.9\\
        pcb & 84.0/81.0 & 89.3/54.5 & \textbf{91.3}/\textbf{87.7} & 86.3/83.8 & \textbf{91.3}/84.9 & 89.2/88.2 & 93.5/66.0 & \textbf{94.6}/\textbf{92.5} & 91.1/90.6 & 94.4/91.5\\
        phone battery & 83.7/83.6 & 86.8/51.6 & 92.5/\textbf{90.6} & 91.6/89.3 & \textbf{93.4}/89.7 & 75.9/80.0 & 81.9/43.8 & 90.7/\textbf{89.1} & 88.6/84.8 & \textbf{94.2}/86.3\\
        plastic nut & 78.7/80.0 & 89.8/59.2 & 91.3/84.9 & 89.4/\textbf{86.2} & \textbf{91.5}/84.8 & 64.7/69.2 & 83.1/40.3 & \textbf{85.7}/77.2 & 82.6/\textbf{79.8} & 85.4/77.5\\
        plastic plug & 70.7/81.4 & 87.5/48.2 & \textbf{89.7}/\textbf{85.2} & 86.1/83.9 & 87.8/84.2 & 59.9/75.9 & 83.7/38.4 & \textbf{87.2}/80.1 & 81.9/\textbf{80.4} & 79.4/78.0\\
        porcelain doll & 68.3/85.1 & 85.4/66.3 & 87.8/\textbf{89.2} & 80.3/85.2 & \textbf{89.8}/85.3 & 53.9/75.2 & 76.8/54.5 & 81.5/\textbf{83.4} & 72.4/79.9 & \textbf{83.3}/76.2\\
        regulator & 46.8/56.9 & 81.7/50.5 & 85.2/66.6 & 80.0/73.6 & \textbf{85.3}/\textbf{73.8} & 26.4/41.5 & 68.3/29.0 & \textbf{75.4}/55.4 & 65.6/\textbf{63.8} & 59.4/58.3\\
        rolled strip base & 97.3/\textbf{98.7} & \textbf{99.5}/59.0 & 99.4/96.9 & 98.8/97.2 & 98.7/98.4 & 98.6/\textbf{99.3} & \textbf{99.7}/75.7 & \textbf{99.7}/98.2 & 99.4/98.5 & 99.2/99.1\\
        sim card set & 91.9/89.7 & 95.2/63.1 & 96.2/94.1 & 93.7/95.1 & \textbf{97.7}/\textbf{96.4} & 90.3/90.3 & 94.8/69.7 & 96.7/94.9 & 94.3/94.8 & \textbf{97.5}/\textbf{96.8}\\
        switch & 89.3/85.5 & \textbf{95.2}/62.2 & 93.1/89.1 & 93.9/\textbf{91.2} & 94.6/88.1 & 91.3/88.6 & \textbf{96.2}/66.8 & 94.8/91.6 & 93.6/\textbf{93.2} & 95.3/91.6\\
        tape & 95.1/\textbf{97.2} & 96.8/49.9 & \textbf{98.1}/96.8 & 97.3/95.6 & 97.5/96.0 & 93.2/\textbf{96.2} & 95.2/41.1 & \textbf{97.4}/96.1 & 93.9/93.2 & 95.6/94.0\\
        terminalblock & 84.4/87.5 & 94.7/59.8 & \textbf{97.3}/93.5 & 97.0/93.0 & 97.2/\textbf{93.7} & 85.8/89.1 & 95.1/64.7 & 97.6/94.4 & 97.5/95.1 & \textbf{97.6}/\textbf{95.4}\\
        toothbrush & 84.9/78.4 & 85.8/65.9 & 85.5/84.8 & 82.1/\textbf{86.5} & \textbf{87.8}/85.7 & 85.4/80.1 & \textbf{87.1}/70.0 & 84.2/86.7 & 86.4/\textbf{87.8} & 86.9/86.1\\
        toy & 79.7/68.4 & 83.5/57.8 & 86.5/79.1 & 85.2/83.9 & \textbf{89.4}/\textbf{85.6} & 82.3/75.1 & 87.6/64.4 & 90.2/84.2 & 85.1/\textbf{87.8} & \textbf{93.3}/87.6\\
        toy brick & 80.0/77.0 & 81.8/58.3 & 77.9/66.4 & 75.6/77.2 & \textbf{83.5}/\textbf{81.6} & 73.9/71.1 & \textbf{78.8}/49.7 & 73.6/58.8 & 72.1/71.7 & 76.9/\textbf{75.5}\\
        transistor1 & 95.8/93.7 & 97.4/62.2 & \textbf{97.9}/94.3 & 96.5/\textbf{94.7} & 93.7/94.6 & 96.6/95.9 & 98.1/69.2 & \textbf{98.4}/96.0 & 94.6/96.1 & 96.6/\textbf{96.2}\\
        u block & 85.4/88.8 & 90.2/62.4 & 93.1/89.1 & 91.8/89.1 & \textbf{93.2}/\textbf{91.5} & 76.7/84.2 & 82.8/48.4 & 90.2/84.0 & 90.3/86.2 & \textbf{93.5}/\textbf{88.0}\\
        usb & 84.5/78.7 & 90.3/57.0 & \textbf{92.8}/\textbf{90.1} & 89.1/88.8 & 92.7/87.9 & 82.9/79.4 & 90.0/55.3 & \textbf{92.1}/\textbf{90.5} & 86.3/88.9 & 91.5/87.6\\
        usb adaptor & 78.3/76.8 & 82.3/47.5 & 83.8/78.1 & 81.7/79.0 & \textbf{87.4}/\textbf{80.2} & 70.3/71.3 & 78.0/38.4 & 78.7/72.4 & 74.1/73.2 & \textbf{82.8}/\textbf{74.4}\\
        vcpill & 83.7/87.1 & 90.3/59.0 & 90.8/83.7 & 89.8/89.0 & \textbf{94.5}/\textbf{90.5} & 81.9/84.0 & 88.8/48.7 & 90.1/80.9 & 88.8/\textbf{87.9} & \textbf{92.7}/87.7\\
        wooden beads & 82.8/78.4 & 86.1/55.1 & 89.5/84.3 & 90.1/86.9 & \textbf{94.6}/\textbf{90.3} & 81.5/77.2 & 84.7/52.0 & 88.9/83.1 & 89.3/86.5 & \textbf{91.3}/\textbf{89.4}\\
        woodstick & 79.7/80.8 & 78.3/58.2 & 85.7/78.0 & 84.1/81.0 & \textbf{86.6}/\textbf{86.0} & 70.4/72.6 & 70.3/35.6 & \textbf{77.9}/65.3 & 74.4/68.9 & 77.0/\textbf{74.4}\\
        zipper & 97.5/98.2 & 98.7/77.2 & \textbf{99.4}/98.9 & 99.1/99.0 & 99.3/\textbf{99.1} & 98.4/98.9 & 99.2/86.7 & \textbf{99.6}/99.4 & 99.4/99.4 & \textbf{99.6}/\textbf{99.5}\\
        \midrule
        Average & 81.6/83.0 & 88.9/57.2 & 90.2/85.2 & 88.1/86.8 & \textbf{91.5}/\textbf{87.8} & 77.3/80.9 & 87.4/53.4 & 88.2/83.2 & 85.6/85.1 & \textbf{88.7}/\textbf{85.3}\\

      \bottomrule
    \end{tabular}
  }
  \label{tab:image_results}
  \vspace{-0.4cm}
\end{table*}

\noindent\textbf{Evaluation Metrics}:
We assess anomaly detection performance at three levels, i.e. image-level, pixel-level, and sample-level. Image-level evaluation measures the model’s ability to distinguish between normal and anomalous instances. {Pixel-level evaluation} quantifies localization accuracy by assessing pixel-wise anomaly scores. {Sample-level evaluation} aggregates anomaly scores from multiple viewpoints of the same sample, computing the maximum score across views (as defined in \cref{eq:inference}), providing a more realistic approximation of industrial inspection scenarios.
For each level, we report Area Under the Receiver Operating Characteristic Curve (AUROC) and Average Precision (AP) as standard performance metrics.

\noindent\textbf{Implementation Details}:
Each image is resized to $224 \times 224$ without data augmentation, and each sample consists of five different viewpoints. The batch size is set to 8, resulting in 40 images per iteration.
We utilize a pretrained DINOv2~\cite{oquab2023dinov2} network as a frozen image encoder, extracting features from the \( 7 \)-th layer of the backbone. DINOv2 employs a $14 \times 14$ patch size and produces feature embeddings of 768 dimensions. For a fair comparison, we replace the baseline backbone of PatchCore~\cite{roth2022towards} (WideResNet50) with the same DINOv2 model.
To train the Epipolar Attention Module, we use the AdamW optimizer~\cite{loshchilov2017decoupled} with a learning rate of $1\text{e-}4$ and a weight decay of $1\text{e-}4$ for 50 epochs. The mask threshold $\delta$ is set to 1 patch, and the number of cluster centers $K$ is set to 20. The $N_k$ in \cref{eq:neg_samp} is set to 1, while the regularization penalty weight $\lambda$ for negative sample-based loss in \cref{eq:final_loss} is set to 0.1.
All experiments are conducted on a single NVIDIA A5000 GPU with 24 GB of memory.

\subsection{Evaluations on Multi-View Industrial Anomaly Detection}

We compared with four state-of-the-art methods. Specifically, UniAD
~\cite{you2022unified} presents a reconstruction based method to simultaneously predict anomaly for all semantic objects. SimpleNet~\cite{liu2023simplenet} adapts pretrained network to industrial data distribution and introduced a generator and discriminator for anomaly detection. MVAD~\cite{he2024learning} proposed to fuse feature across multiple views via similarity based cross-view attention. PatchCore~\cite{roth2022towards} is a memory bank based anomaly detection, we extract feature on each view separately for PatchCore.

\noindent\textbf{Sample-Level Evaluation}: We first present the evaluations on sample-level anomaly detection for both single-class and multi-class protocols in \cref{tab:sample_results}. We make the following observations from the results. 
i) Superior Performance of MVEAD: Our proposed MVEAD consistently outperforms all competing methods across both AUROC and AP metrics. Under the single-class protocol, MVEAD improves upon the state-of-the-art by 1.6\% in AUROC and 0.5\% in AP.
ii) Greater Gains in the Multi-Class Setting: Under the more challenging multi-class protocol, MVEAD achieves an even more pronounced improvement, with 1.8\% higher AUROC. Notably, strong baselines such as SimpleNet (94.6\% $\rightarrow$ 60.8\% AUROC) and MVAD (94.3\% $\rightarrow$ 89.0\% AUROC) suffer significant performance degradation in this setting, likely due to confusion between different object classes. While UniAD shows slightly better performance in the multi-class protocol, its overall performance lags behind other competitors.
iii) Effectiveness of Memory Bank-Based Methods: Methods based on memory banks, such as PatchCore and MVEAD, demonstrate a stronger ability to capture the complex distribution of normal samples, thereby maintaining better performance in the multi-class setting.


\noindent\textbf{Image-Level Evaluation}: We further present the evaluations at image-level anomaly detection in \cref{tab:image_results} with the following key findings.
i) Image-Level Detection is More Challenging: Compared to sample-level evaluations, image-level anomaly detection generally yields lower AUROC and AP scores, as defects may be less noticeable from certain views, making anomaly identification more difficult.
ii) Cross-View Attention Benefits Image-Level Detection: While PatchCore outperforms MVAD at the sample level, MVAD consistently surpasses PatchCore in image-level evaluation, highlighting the benefits of cross-view feature fusion. However, our MVEAD still achieves the best performance across all metrics. The significant improvement over PatchCore (88.1\% $\rightarrow$ 91.5\% AUROC), which shares a similar inference pipeline, underscores the effectiveness of our epipolar attention module.
iii) Lower AP Scores Due to Class Imbalance: We observe relatively lower AP scores in image-level evaluations, likely due to the higher class imbalance—many views may not contain visible anomalies. This suggests that AP should be given more consideration in highly multi-view settings, where class imbalance is more severe than in single-view scenarios.


\noindent\textbf{Qualitative Evaluation}: We present qualitative evaluations in \cref{fig:vis_sample}. In general, our method is able to better localize the correct anomalous region. In comparison, UniAD produces many false positive predictions forming many local maximums which could severely confuse human understanding. Compared with PatchCore, MVEAD is able to better localize the defects. However, there are some blurry ghost predictions in MVEAD. This might be caused by accidentally fusing anomalous features from other views. This suggests potential areas for improvement in refining cross-view attention mechanisms to mitigate feature leakage.

\begin{table}
  \centering
   \caption{Ablation study on cross-view fusion methods and pre-training strategies, evaluated by Image-AUROC (\%).}
     \vspace{-0.2cm}
     \setlength{\tabcolsep}{3pt}
  \resizebox{0.8\linewidth}{!}{
  \begin{tabular}{@{}lccc@{}}
    \toprule
    Cross-View Attn. & Pretrain. & MultiView MB & Results \\
    \midrule
    -        & -               & -  & 88.1
\\
    MVAS~\cite{he2024learning}   & Copy               & -  & 82.5\\
    MVAS~\cite{he2024learning}   & DeepSVDD~\cite{ruff2018deep}           & -  & 85.9\\
    MVAS~\cite{he2024learning}   & MCP       & -  & 87.7\\
    MVAS~\cite{he2024learning}   & MCP+Reg   & -  & 88.6 \\
    EAM     & Copy               & -  & 84.9\\
    EAM     & DeepSVDD~\cite{ruff2018deep}           & -  & 86.8 \\
    EAM     & MCP  & -  & 89.6 \\
    EAM     & MCP+Reg  & -  & 90.9 \\
    EAM     & MCP+Reg  & $\checkmark$   & 91.5 \\
    \bottomrule
  \end{tabular}
  }
  \label{tab:main_ablation}
  \vspace{-0.4cm}
\end{table}

\noindent\textbf{Ablation Study}: We investigate two types of cross-view fusion methods~(Cross-View Attn.), including semantic similarity based multi-view adaptive selection~(\textbf{MVAS}) proposed by~\cite{he2024learning} and our epipolar attention module~(\textbf{EAM}). We further investigate different pretraining strategies. These include directly copying the projection layer from the \( 8 \)-th layer of the backbone which adopts to the output of the \( 7 \)-th layer in the backbone~(\textbf{Copy}), reducing volume in feature space(DeepSVDD~\cite{ruff2018deep}), our proposed multi-center pretraining~(\textbf{MCP}) and additionally included the negative samples for regularization~(\textbf{MCP}+\textbf{Reg}). Finally, we also include the multiview memory bank~(\textbf{MultiView MB}).

We make the following observations from the results in \cref{tab:main_ablation}. i) Cross-view attention must be properly trained to be effective. For both MVAS and EAM, directly copying the weights yields inferior results than without any cross-view fusion. ii) Reducing data volume in the features space~(DeepSVDD) provides signal to train the projection layers and improves upon without training, however is still behind the baseline. iii) Multi-center pretraining can further boost the performance, suggesting encouraging local clusters in feature space is conducive to memory bank based anomaly inference. iv) Both multi-view memory bank and negative sample synthesis as regularization are complementary to the overall method.

\begin{figure}[h]
  \centering
  \begin{subfigure}{\linewidth}
    \centering
    \includegraphics[width=1.03\linewidth]{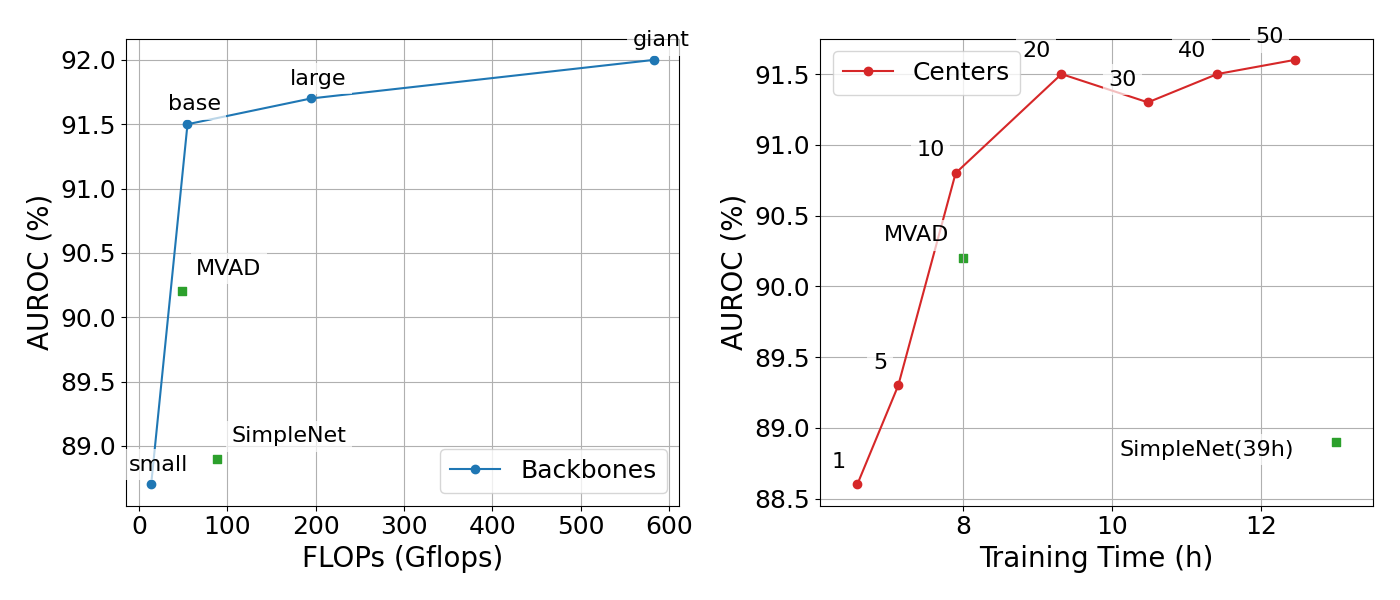}
    \label{fig:short-a}
  \end{subfigure}
  
  \vspace{-0.4cm}
  \caption{Computation overhead for different backbone sizes (DINOv2-small/base/large/giant) and center numbers $K$.}
  \label{fig:backbone_center}
  
  \vspace{-0.4cm}
\end{figure}

\subsection{Additional Studies}

We demonstrate that overall performance remains relatively unaffected by computational overhead. As shown in the left subfigure of \cref{fig:backbone_center}, DINOv2-small exhibits low computational cost but achieves only 88.7\% AUROC, suggesting that its limited hidden dimensions fail to capture fine-grained anomalies. In contrast, DINOv2-giant attains the highest AUROC (92.0\%) but incurs a substantial computational burden (over 600 GFlops). The DINOv2 base model strikes a favorable balance, achieving 91.5\% AUROC while maintaining significantly lower FLOPs than its larger counterpart. In comparison, SimpleNet and MVAD are both below the Pareto frontier established by our method.

We also analyze the impact of the number of cluster centers $K$ in $\set{M}$ on performance and inference overhead. As shown in the right subfigure of \cref{fig:backbone_center}, increasing $K$ generally improves AUROC but extends training time. A single center yields an AUROC of 88.6\% with the shortest training time (6.58 hours), while increasing $K$ to 20 boosts AUROC to 91.5\% at the cost of a longer training duration (9.32 hours). However, beyond $K=20$, AUROC saturates despite the rising computational burden, indicating that 20 centers sufficiently model the patch feature distribution. In comparison, MVAD incurs a similar training cost but yields inferior performance, while SimpleNet requires five times the training cost and performs even worse.

\section{Conclusion}


We explored improving the efficacy of multi-view anomaly detection by introducing cross-view attention guided by epipolar geometry. Using the fundamental matrix, we localize potential patches in the support view based on a given patch in the reference view. A cross-view attention (EAM) is defined only on the potential patches to eliminate the impact from irrelevant ones. Inspired by the memory bank-based anomaly detection method, we propose a multi-center pre-training strategy. Negative samples are synthesized to regularize the pre-training. Evaluations on Real-IAD dataset demonstrate the effectiveness, achieving superior performance in both single-view and sample-wise anomaly prediction. 

{
    \small
    \bibliographystyle{ieeenat_fullname}
    \bibliography{main}

\begin{thebibliography}{45}
\providecommand{\natexlab}[1]{#1}
\providecommand{\url}[1]{\texttt{#1}}
\expandafter\ifx\csname urlstyle\endcsname\relax
  \providecommand{\doi}[1]{doi: #1}\else
  \providecommand{\doi}{doi: \begingroup \urlstyle{rm}\Url}\fi

\bibitem[Liu et~al.(2024)Liu, Xie, Wang, Li, Wang, Zheng, and Jin]{liu2024deep}
Jiaqi Liu, Guoyang Xie, Jinbao Wang, Shangnian Li, Chengjie Wang, Feng Zheng, and Yaochu Jin.
\newblock Deep industrial image anomaly detection: A survey.
\newblock \emph{Machine Intelligence Research}, 2024.

\bibitem[Xie et~al.(2024)Xie, Wang, Liu, Lyu, Liu, Wang, Zheng, and Jin]{xie2024iad}
Guoyang Xie, Jinbao Wang, Jiaqi Liu, Jiayi Lyu, Yong Liu, Chengjie Wang, Feng Zheng, and Yaochu Jin.
\newblock Im-iad: Industrial image anomaly detection benchmark in manufacturing.
\newblock \emph{IEEE Transactions on Cybernetics}, 2024.

\bibitem[Bergmann et~al.(2019)Bergmann, Fauser, Sattlegger, and Steger]{bergmann2019mvtec}
Paul Bergmann, Michael Fauser, David Sattlegger, and Carsten Steger.
\newblock Mvtec ad--a comprehensive real-world dataset for unsupervised anomaly detection.
\newblock In \emph{IEEE/CVF Conference on Computer Vision and Pattern Recognition}, 2019.

\bibitem[Zou et~al.(2022)Zou, Jeong, Pemula, Zhang, and Dabeer]{zou2022spot}
Yang Zou, Jongheon Jeong, Latha Pemula, Dongqing Zhang, and Onkar Dabeer.
\newblock Spot-the-difference self-supervised pre-training for anomaly detection and segmentation.
\newblock In \emph{European Conference on Computer Vision}, 2022.

\bibitem[Wang et~al.(2024)Wang, Zhu, Gao, Gan, Zhang, Gu, Qian, Chen, and Ma]{wang2024real}
Chengjie Wang, Wenbing Zhu, Bin-Bin Gao, Zhenye Gan, Jiangning Zhang, Zhihao Gu, Shuguang Qian, Mingang Chen, and Lizhuang Ma.
\newblock Real-iad: A real-world multi-view dataset for benchmarking versatile industrial anomaly detection.
\newblock In \emph{IEEE/CVF Conference on Computer Vision and Pattern Recognition}, 2024.

\bibitem[Wang et~al.(2022)Wang, Liu, Yu, Liu, Zhou, Zhu, Yang, Yin, and Yang]{wang2022multiview}
Siqi Wang, Jiyuan Liu, Guang Yu, Xinwang Liu, Sihang Zhou, En Zhu, Yuexiang Yang, Jianping Yin, and Wenjing Yang.
\newblock Multiview deep anomaly detection: A systematic exploration.
\newblock \emph{IEEE Transactions on Neural Networks and Learning Systems}, 2022.

\bibitem[Wang et~al.(2018)Wang, Girshick, Gupta, and He]{wang2018non}
Xiaolong Wang, Ross Girshick, Abhinav Gupta, and Kaiming He.
\newblock Non-local neural networks.
\newblock In \emph{IEEE/CVF Conference on Computer Vision and Pattern Recognition}, 2018.

\bibitem[He et~al.(2024)He, Zhang, Tian, Wang, and Xie]{he2024learning}
Haoyang He, Jiangning Zhang, Guanzhong Tian, Chengjie Wang, and Lei Xie.
\newblock Learning multi-view anomaly detection.
\newblock \emph{arXiv preprint arXiv:2407.11935}, 2024.

\bibitem[Deng and Li(2022)]{deng2022anomaly}
Hanqiu Deng and Xingyu Li.
\newblock Anomaly detection via reverse distillation from one-class embedding.
\newblock In \emph{IEEE/CVF Conference on Computer Vision and Pattern Recognition}, 2022.

\bibitem[Hartley and Zisserman(2003)]{hartley2003multiple}
Richard Hartley and Andrew Zisserman.
\newblock \emph{Multiple view geometry in computer vision}.
\newblock Cambridge university press, 2003.

\bibitem[Roth et~al.(2022)Roth, Pemula, Zepeda, Sch{\"o}lkopf, Brox, and Gehler]{roth2022towards}
Karsten Roth, Latha Pemula, Joaquin Zepeda, Bernhard Sch{\"o}lkopf, Thomas Brox, and Peter Gehler.
\newblock Towards total recall in industrial anomaly detection.
\newblock In \emph{IEEE/CVF Conference on Computer Vision and Pattern Recognition}, 2022.

\bibitem[Tien et~al.(2023)Tien, Nguyen, Tran, Huy, Duong, Nguyen, and Truong]{tien2023revisiting}
Tran~Dinh Tien, Anh~Tuan Nguyen, Nguyen~Hoang Tran, Ta~Duc Huy, Soan Duong, Chanh D~Tr Nguyen, and Steven~QH Truong.
\newblock Revisiting reverse distillation for anomaly detection.
\newblock In \emph{IEEE/CVF Conference on Computer Vision and Pattern Recognition}, 2023.

\bibitem[He et~al.(2020)He, Yan, Fragkiadaki, and Yu]{he2020epipolar}
Yihui He, Rui Yan, Katerina Fragkiadaki, and Shoou-I Yu.
\newblock Epipolar transformers.
\newblock In \emph{IEEE/CVF Conference on Computer Vision and Pattern Recognition}, 2020.

\bibitem[Ruff et~al.(2018)Ruff, Vandermeulen, Goernitz, Deecke, Siddiqui, Binder, M{\"u}ller, and Kloft]{ruff2018deep}
Lukas Ruff, Robert Vandermeulen, Nico Goernitz, Lucas Deecke, Shoaib~Ahmed Siddiqui, Alexander Binder, Emmanuel M{\"u}ller, and Marius Kloft.
\newblock Deep one-class classification.
\newblock In \emph{International Conference on Machine Learning}, 2018.

\bibitem[Li et~al.(2021)Li, Sohn, Yoon, and Pfister]{li2021cutpaste}
Chun-Liang Li, Kihyuk Sohn, Jinsung Yoon, and Tomas Pfister.
\newblock Cutpaste: Self-supervised learning for anomaly detection and localization.
\newblock In \emph{IEEE/CVF Conference on Computer Vision and Pattern Recognition}, 2021.

\bibitem[Zhang et~al.(2024)Zhang, Xu, and Zhou]{zhang2024realnet}
Ximiao Zhang, Min Xu, and Xiuzhuang Zhou.
\newblock Realnet: A feature selection network with realistic synthetic anomaly for anomaly detection.
\newblock In \emph{IEEE/CVF Conference on Computer Vision and Pattern Recognition}, 2024.

\bibitem[Cao et~al.(2024)Cao, Xu, Zhang, Cheng, Huang, Pang, and Shen]{cao2024survey}
Yunkang Cao, Xiaohao Xu, Jiangning Zhang, Yuqi Cheng, Xiaonan Huang, Guansong Pang, and Weiming Shen.
\newblock A survey on visual anomaly detection: Challenge, approach, and prospect.
\newblock \emph{arXiv preprint arXiv:2401.16402}, 2024.

\bibitem[Liu et~al.(2023)Liu, Zhou, Xu, and Wang]{liu2023simplenet}
Zhikang Liu, Yiming Zhou, Yuansheng Xu, and Zilei Wang.
\newblock Simplenet: A simple network for image anomaly detection and localization.
\newblock In \emph{IEEE/CVF Conference on Computer Vision and Pattern Recognition}, 2023.

\bibitem[Zavrtanik et~al.(2021)Zavrtanik, Kristan, and Sko{\v{c}}aj]{zavrtanik2021draem}
Vitjan Zavrtanik, Matej Kristan, and Danijel Sko{\v{c}}aj.
\newblock Draem-a discriminatively trained reconstruction embedding for surface anomaly detection.
\newblock In \emph{IEEE/CVF International Conference on Computer Vision}, 2021.

\bibitem[Cimpoi et~al.(2014)Cimpoi, Maji, Kokkinos, Mohamed, and Vedaldi]{cimpoi2014describing}
Mircea Cimpoi, Subhransu Maji, Iasonas Kokkinos, Sammy Mohamed, and Andrea Vedaldi.
\newblock Describing textures in the wild.
\newblock In \emph{IEEE/CVF Conference on Computer Vision and Pattern Recognition}, 2014.

\bibitem[Ho et~al.(2020)Ho, Jain, and Abbeel]{ho2020denoising}
Jonathan Ho, Ajay Jain, and Pieter Abbeel.
\newblock Denoising diffusion probabilistic models.
\newblock \emph{Advances in Neural Information Processing Systems}, 2020.

\bibitem[Akcay et~al.(2019)Akcay, Atapour-Abarghouei, and Breckon]{akcay2019ganomaly}
Samet Akcay, Amir Atapour-Abarghouei, and Toby~P Breckon.
\newblock Ganomaly: Semi-supervised anomaly detection via adversarial training.
\newblock 2019.

\bibitem[Schlegl et~al.(2019)Schlegl, Seeb{\"o}ck, Waldstein, Langs, and Schmidt-Erfurth]{schlegl2019f}
Thomas Schlegl, Philipp Seeb{\"o}ck, Sebastian~M Waldstein, Georg Langs, and Ursula Schmidt-Erfurth.
\newblock f-anogan: Fast unsupervised anomaly detection with generative adversarial networks.
\newblock \emph{Medical image analysis}, 2019.

\bibitem[An and Cho(2015)]{an2015variational}
Jinwon An and Sungzoon Cho.
\newblock Variational autoencoder based anomaly detection using reconstruction probability.
\newblock \emph{Special lecture on IE}, 2015.

\bibitem[Zavrtanik et~al.(2021)Zavrtanik, Kristan, and Sko{\v{c}}aj]{zavrtanik2021reconstruction}
Vitjan Zavrtanik, Matej Kristan, and Danijel Sko{\v{c}}aj.
\newblock Reconstruction by inpainting for visual anomaly detection.
\newblock \emph{Pattern Recognition}, 2021.

\bibitem[Collin and De~Vleeschouwer(2021)]{collin2021improved}
Anne-Sophie Collin and Christophe De~Vleeschouwer.
\newblock Improved anomaly detection by training an autoencoder with skip connections on images corrupted with stain-shaped noise.
\newblock 2021.

\bibitem[Liao et~al.(2024)Liao, Xu, Nguyen, Goodge, and Foo]{liao2024coft}
Jingyi Liao, Xun Xu, Manh~Cuong Nguyen, Adam Goodge, and Chuan~Sheng Foo.
\newblock Coft-ad: Contrastive fine-tuning for few-shot anomaly detection.
\newblock \emph{IEEE Transactions on Image Processing}, 2024.

\bibitem[Defard et~al.(2021)Defard, Setkov, Loesch, and Audigier]{defard2021padim}
Thomas Defard, Aleksandr Setkov, Angelique Loesch, and Romaric Audigier.
\newblock Padim: a patch distribution modeling framework for anomaly detection and localization.
\newblock 2021.

\bibitem[Kobyzev et~al.(2020)Kobyzev, Prince, and Brubaker]{kobyzev2020normalizing}
Ivan Kobyzev, Simon~JD Prince, and Marcus~A Brubaker.
\newblock Normalizing flows: An introduction and review of current methods.
\newblock \emph{IEEE Transactions on Pattern analysis and Machine Intelligence}, 2020.

\bibitem[Rudolph et~al.(2021)Rudolph, Wandt, and Rosenhahn]{rudolph2021same}
Marco Rudolph, Bastian Wandt, and Bodo Rosenhahn.
\newblock Same same but differnet: Semi-supervised defect detection with normalizing flows.
\newblock In \emph{IEEE/CVF Winter Conference on Applications of Computer Vision}, 2021.

\bibitem[Su et~al.(2015)Su, Maji, Kalogerakis, and Learned-Miller]{su2015multi}
Hang Su, Subhransu Maji, Evangelos Kalogerakis, and Erik Learned-Miller.
\newblock Multi-view convolutional neural networks for 3d shape recognition.
\newblock In \emph{IEEE International Conference on Computer Vision}, 2015.

\bibitem[Qiu et~al.(2019)Qiu, Wang, Wang, Wang, and Zeng]{qiu2019cross}
Haibo Qiu, Chunyu Wang, Jingdong Wang, Naiyan Wang, and Wenjun Zeng.
\newblock Cross view fusion for 3d human pose estimation.
\newblock In \emph{IEEE/CVF International Conference on Computer Vision}, 2019.

\bibitem[Ma et~al.(2022)Ma, Wang, Chen, Kong, Chen, Liu, Yan, Tang, and Xie]{ma2022ppt}
Haoyu Ma, Zhe Wang, Yifei Chen, Deying Kong, Liangjian Chen, Xingwei Liu, Xiangyi Yan, Hao Tang, and Xiaohui Xie.
\newblock Ppt: token-pruned pose transformer for monocular and multi-view human pose estimation.
\newblock In \emph{European Conference on Computer Vision}, 2022.

\bibitem[Zhang et~al.(2021)Zhang, Wang, Qiu, Qin, and Zeng]{zhang2021adafuse}
Zhe Zhang, Chunyu Wang, Weichao Qiu, Wenhu Qin, and Wenjun Zeng.
\newblock Adafuse: Adaptive multiview fusion for accurate human pose estimation in the wild.
\newblock \emph{International Journal of Computer Vision}, 2021.

\bibitem[Furukawa et~al.(2015)Furukawa, Hern{\'a}ndez, et~al.]{furukawa2015multi}
Yasutaka Furukawa, Carlos Hern{\'a}ndez, et~al.
\newblock Multi-view stereo: A tutorial.
\newblock \emph{Foundations and trends{\textregistered} in Computer Graphics and Vision}, 2015.

\bibitem[Gu et~al.(2020)Gu, Fan, Zhu, Dai, Tan, and Tan]{gu2020cascade}
Xiaodong Gu, Zhiwen Fan, Siyu Zhu, Zuozhuo Dai, Feitong Tan, and Ping Tan.
\newblock Cascade cost volume for high-resolution multi-view stereo and stereo matching.
\newblock In \emph{IEEE/CVF Conference on Computer Vision and Pattern Recognition}, 2020.

\bibitem[Sheng et~al.(2019)Sheng, Zhan, Lu, and Jiang]{sheng2019multi}
Xiang-Rong Sheng, De-Chuan Zhan, Su Lu, and Yuan Jiang.
\newblock Multi-view anomaly detection: Neighborhood in locality matters.
\newblock In \emph{AAAI Conference on Artificial Intelligence}, 2019.

\bibitem[Xu and Tao(2020)]{xu2020pvsnet}
Qingshan Xu and Wenbing Tao.
\newblock Pvsnet: Pixelwise visibility-aware multi-view stereo network.
\newblock \emph{arXiv preprint arXiv:2007.07714}, 2020.

\bibitem[Wang et~al.(2021)Wang, Galliani, Vogel, Speciale, and Pollefeys]{wang2021patchmatchnet}
Fangjinhua Wang, Silvano Galliani, Christoph Vogel, Pablo Speciale, and Marc Pollefeys.
\newblock Patchmatchnet: Learned multi-view patchmatch stereo.
\newblock In \emph{IEEE/CVF Conference on Computer Vision and Pattern Recognition}, 2021.

\bibitem[Wang and Lan(2020)]{wang2020towards}
Zhen Wang and Chao Lan.
\newblock Towards a hierarchical bayesian model of multi-view anomaly detection.
\newblock In \emph{International Joint Conference on Artificial Intelligence}, 2020.

\bibitem[Liu et~al.(2024)Liu, Chu, Hsieh, Chen, and Liu]{liu2024learning}
Chieh Liu, Yu-Min Chu, Ting-I Hsieh, Hwann-Tzong Chen, and Tyng-Luh Liu.
\newblock Learning diffusion models for multi-view anomaly detection.
\newblock In \emph{European Conference on Computer Vision}, 2024.

\bibitem[Dosovitskiy et~al.(2020)Dosovitskiy, Beyer, Kolesnikov, Weissenborn, Zhai, Unterthiner, Dehghani, Minderer, Heigold, Gelly, et~al.]{dosovitskiy2020image}
Alexey Dosovitskiy, Lucas Beyer, Alexander Kolesnikov, Dirk Weissenborn, Xiaohua Zhai, Thomas Unterthiner, Mostafa Dehghani, Matthias Minderer, Georg Heigold, Sylvain Gelly, et~al.
\newblock An image is worth 16x16 words: Transformers for image recognition at scale.
\newblock \emph{arXiv preprint arXiv:2010.11929}, 2020.

\bibitem[You et~al.(2022)You, Cui, Shen, Yang, Lu, Zheng, and Le]{you2022unified}
Zhiyuan You, Lei Cui, Yujun Shen, Kai Yang, Xin Lu, Yu Zheng, and Xinyi Le.
\newblock A unified model for multi-class anomaly detection.
\newblock \emph{Advances in Neural Information Processing Systems}, 2022.

\bibitem[Oquab et~al.(2023)Oquab, Darcet, Moutakanni, Vo, Szafraniec, Khalidov, Fernandez, Haziza, Massa, El-Nouby, et~al.]{oquab2023dinov2}
Maxime Oquab, Timoth{\'e}e Darcet, Th{\'e}o Moutakanni, Huy Vo, Marc Szafraniec, Vasil Khalidov, Pierre Fernandez, Daniel Haziza, Francisco Massa, Alaaeldin El-Nouby, et~al.
\newblock Dinov2: Learning robust visual features without supervision.
\newblock \emph{arXiv preprint arXiv:2304.07193}, 2023.

\bibitem[Loshchilov and Hutter(2017)]{loshchilov2017decoupled}
Ilya Loshchilov and Frank Hutter.
\newblock Decoupled weight decay regularization.
\newblock \emph{arXiv preprint arXiv:1711.05101}, 2017.

\end{thebibliography}
}


\end{document}


\clearpage
\setcounter{page}{1}
\maketitlesupplementary
\section*{A. Additional Implementation Details}

For the baseline methods, including UniAD\cite{you2022unified}, SimpleNet\cite{liu2023simplenet}, and MVAD\cite{he2024learning}, we follow the experimental settings and results provided in MVAD. For PatchCore\cite{roth2022towards}, to ensure consistency with our method, we replace the default WideResNet50 backbone with DINOv2\cite{oquab2023dinov2} and use the output from the 7th layer to construct the memory bank. All other hyperparameters remain the same as in PatchCore.

In the multi-class setting, both our method and PatchCore reduce the coreset downsampling ratio for the memory bank from 10\% to 0.33\% to ensure a consistent memory bank size.

\section*{B. Additional Study}

\subsection*{B.1 Feature Selection}
\begin{figure}[h]
  \centering
  \begin{subfigure}{\linewidth}
    \centering
    \includegraphics[width=\linewidth]{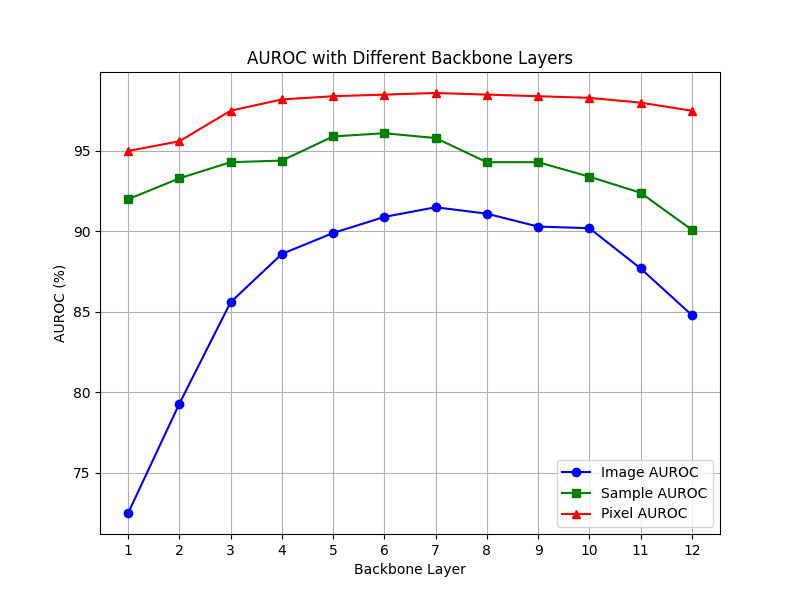}
    \label{fig:short-a}
  \end{subfigure}
  \caption{Result(Image-AUROC\%) on different backbone layers}
  \label{fig:layer_fig}
\end{figure}

We investigate the impact of selecting different block feature for anomaly detection.
\cref{fig:layer_fig} illustrates the effect of using outputs from different hidden layers of the backbone as inputs to the EAM module. We find that with DINOv2-base (12-layer), the 7th layer yields the best results. This contrasts with CNN-based architectures like ResNet, where earlier layers (e.g., \( 2 \)-nd or \( 2 \)-rd) typically perform best. The difference stems from transformers' gradual accumulation of global contextual information across layers, whereas CNNs focus on local feature extraction in the early stages. 

\cref{tab:layer_tab} further examines the results obtained by combining different layers near the optimal selection. Combining the outputs of the \( 6 \)-th or \( 8 \)-th layer with the \( 7 \)-th layer also achieves comparable results but slightly increases computational cost. Therefore, we use only a single-layer output as the input to the EAM module.

\subsection*{B.2 Epipolar Attention Thresholding}
\begin{figure}[h]
  \centering
  \begin{subfigure}{\linewidth}
    \centering
    \includegraphics[width=\linewidth]{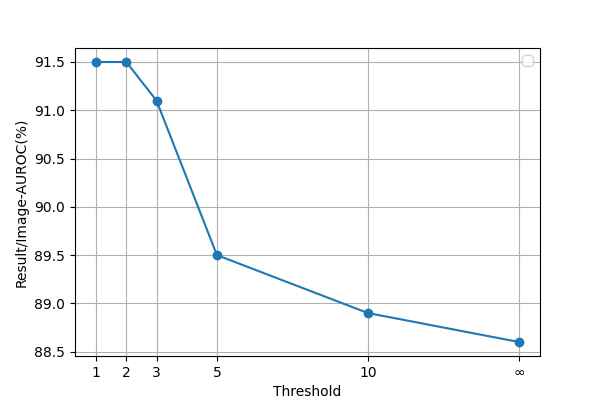}
    \label{fig:short-a}
  \end{subfigure}
  \caption{Result(Image-AUROC\%) on different threshold $\delta$~(patch)}
  \label{fig:threshold}
\end{figure}

We use a threshold $\delta$ to select the patches for cross-view attention.
\cref{fig:threshold} illustrates the effect of the epipolar mask size used in the EAM (Epipolar Attention Module) and the threshold $\delta$ in Eq. (11) in Sec. (3.3) on the final results.
The ideal epipolar line should be a one-pixel-wide straight line. However, since the feature map comes from the intermediate layers of a ViT-like backbone, the representation of the epipolar line can only be achieved at a patchwise level. To minimize the visible region for each pixel, we set the threshold to 1 patch. As the threshold increases, allowing a larger visible space, the performance correspondingly decreases. When the threshold approaches infinity, EAM degenerates into standard Cross-Attention, yielding the lowest performance. This confirms the effectiveness of the epipolar mask in the EAM module.

\subsection*{B.3 Distribution of Prototypes in Memory Bank}
\begin{figure*}[h]
  \centering
  \begin{subfigure}{\linewidth}
    \centering
    \includegraphics[width=\linewidth]{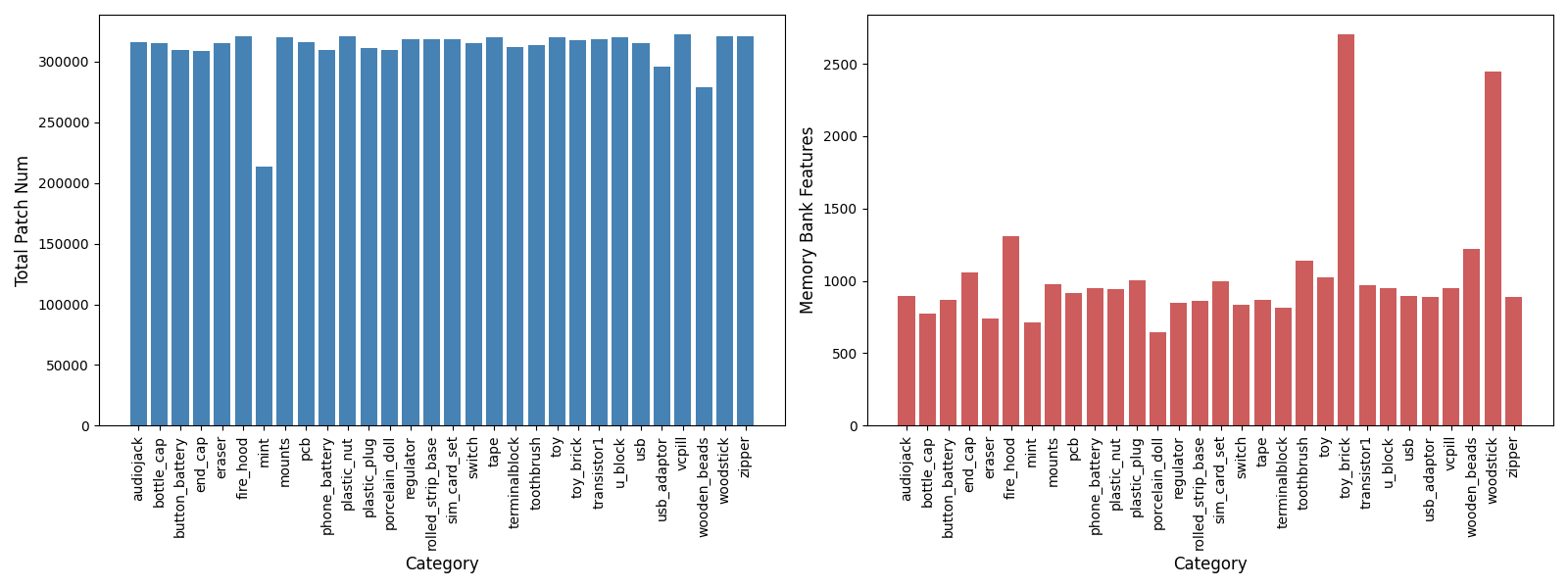}
    \label{fig:short-a}
  \end{subfigure}
  \caption{Category distribution on Real-IAD dataset and downsampled Multi-class memory bank}
  \label{fig:bank_histogram}
\end{figure*}

\cref{fig:bank_histogram} illustrates the proportion of each subclass in the Memory Bank under the multi-class setting. The left figure shows the distribution of the number of patches extracted from each subclass, while the right figure presents the number of patches retained in the Memory Bank after core-set selection. Our method achieves a relatively balanced class distribution. The higher proportion of the two prominent subclasses, toy brick and wood stick, is due to their richer texture details compared to industrial components, leading to a more complex patch feature space.

\subsection*{B.4 Impact of Synthetic Negative Sample Regularization}

\begin{figure}[h]
  \centering
  \begin{subfigure}{\linewidth}
    \centering
    \includegraphics[width=\linewidth]{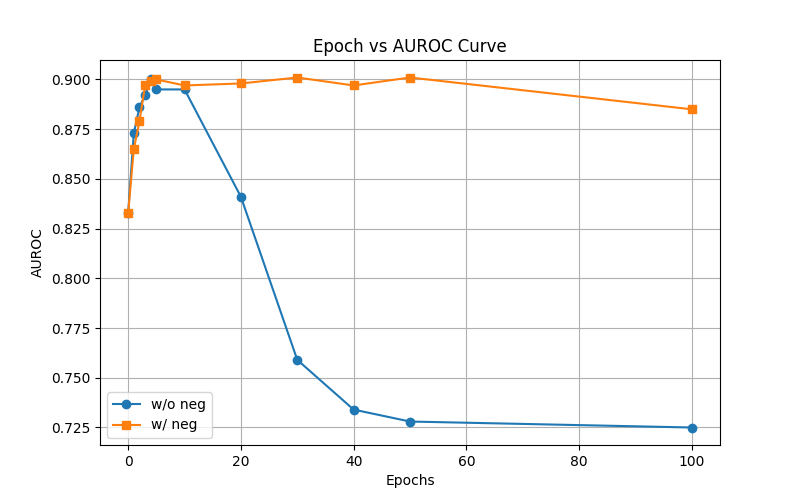}
    \label{fig:short-a}
  \end{subfigure}
  \caption{Result(Image-AUROC\%) on different epoches with/without negative sample regularization on audiojack subclass}
  \label{fig:neg_samp}
\end{figure}

We study the impact of negative sample regularization in Fig.~\ref{fig:neg_samp} by examining the AUROC on testing set. It is clearly observed that if synthetic negative samples are not introduced as regularization, the pre-training has a high risk of collapsing after 10 training epochs. In contrast, synthetic negative samples can mitigate the model collapsing issue.

\section*{C. Visualization on Real-IAD dataset}

In this section, we provide visualization results of our method's anomaly segmentation across 30 subclasses over Real-IAD dataset.

\begin{table}[h]
    \centering
    \begin{tabular}{cccc}
        \hline
        6 & 7 & 8 & Image-AUROC/\% \\
        \hline
        $\checkmark$ & - & - & 90.9 \\
        - &$ \checkmark$ & - & 91.5 \\
        - & - & $\checkmark$ & 91.1 \\
        $\checkmark$ & $\checkmark$ & 0 & 91.5 \\
        - & $\checkmark$ & $\checkmark$ & 91.3 \\
        $\checkmark$ & $\checkmark$ & $\checkmark$ & 91.0 \\
        \hline
    \end{tabular}
    \caption{Experimental Results for Different Layer Configurations}
    \label{tab:layer_tab}
\end{table}

\section*{C. Visualization on Real-IAD dataset}
\begin{figure*}[h]
  \centering
  \begin{subfigure}{\linewidth}
    \centering
    \includegraphics[width=\linewidth]{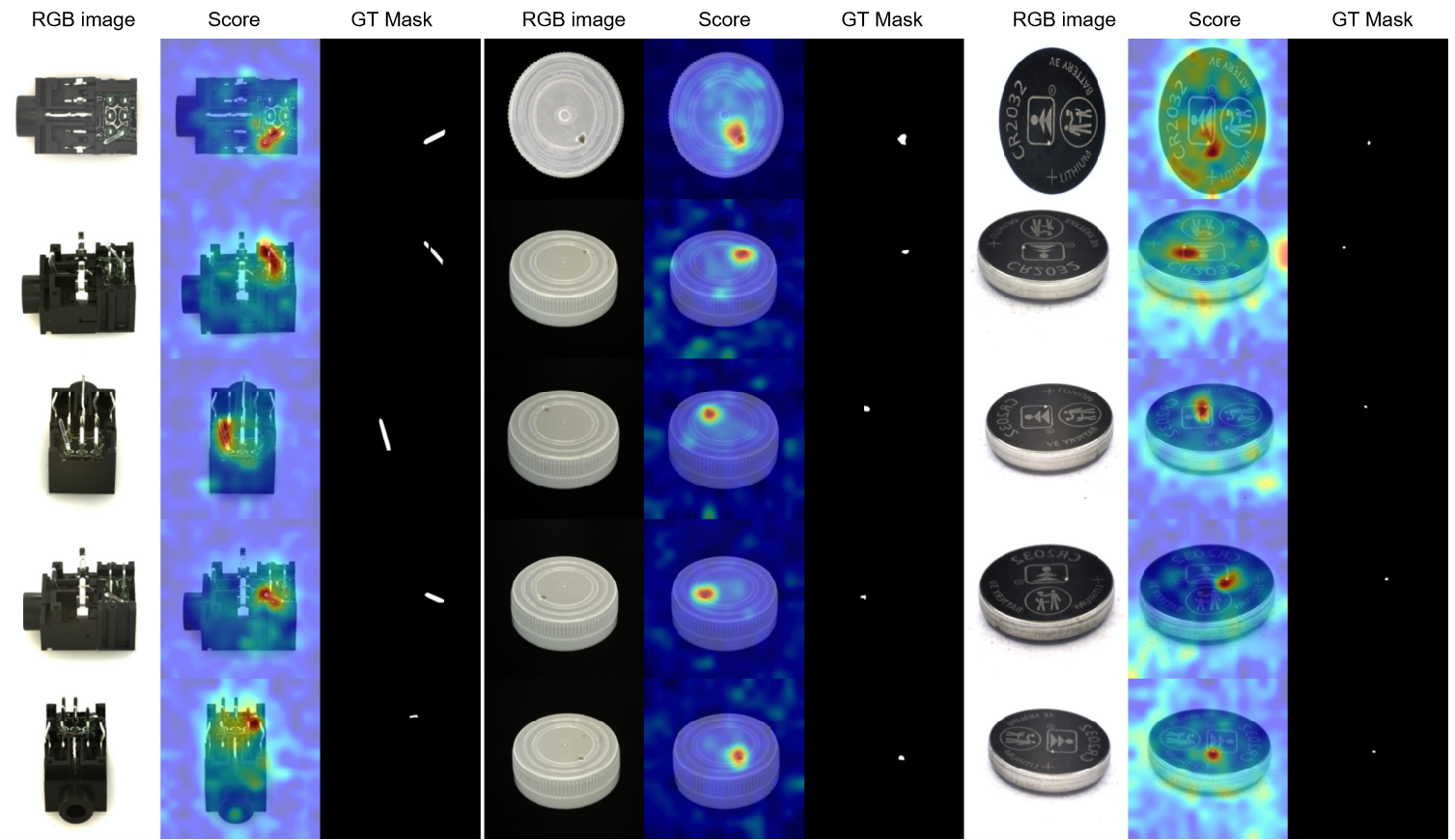}
    \label{fig:short-a}
  \end{subfigure}
  \caption{Visualization of our method on audiojack, bottle cap, and button battery (from left to right).}
  \label{fig:full_vis_1}
\end{figure*}

\begin{figure*}[h]
  \centering
  \begin{subfigure}{\linewidth}
    \centering
    \includegraphics[width=\linewidth]{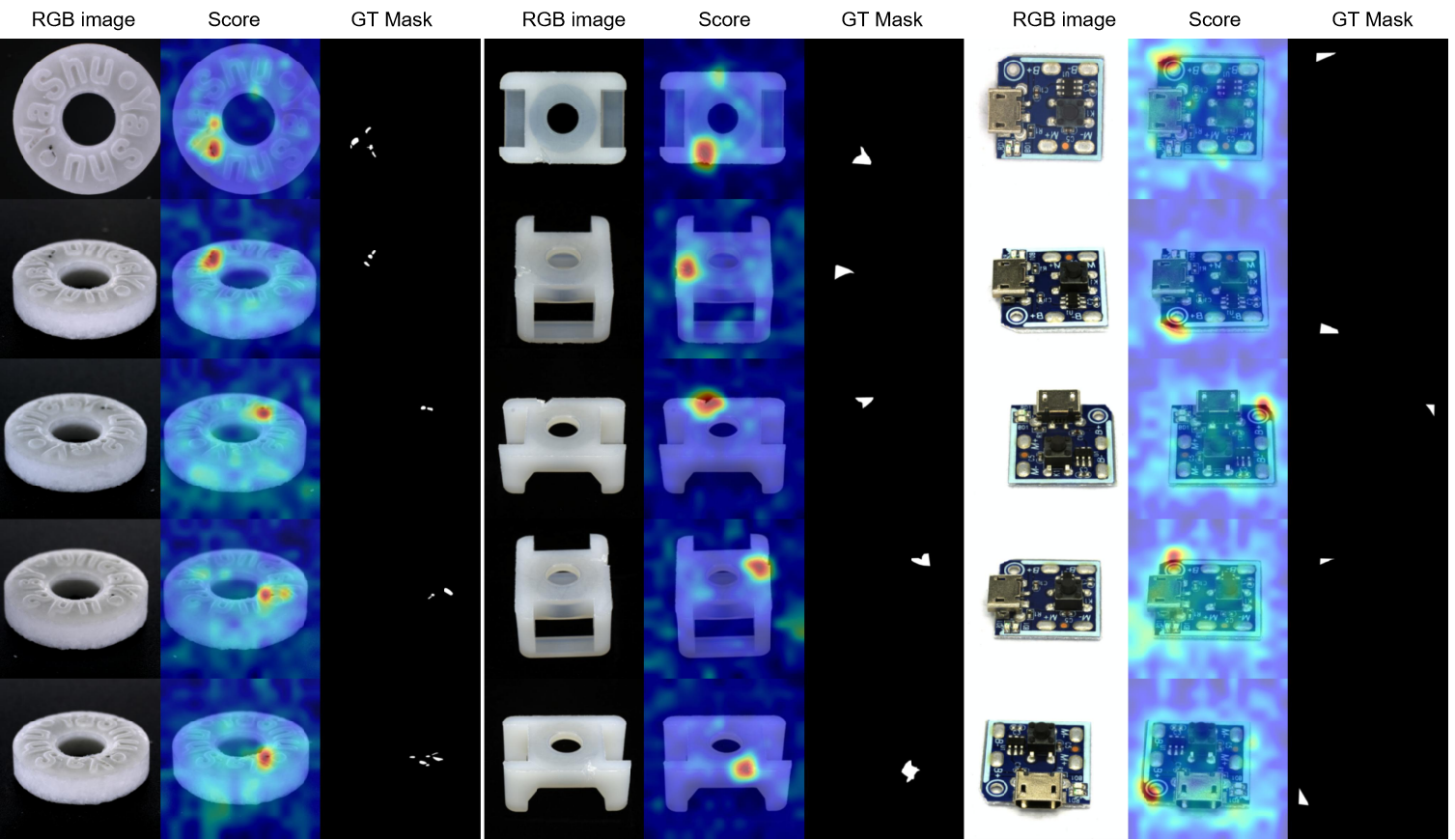}
    \label{fig:short-a}
  \end{subfigure}
  \caption{Visualization of our method on end cap, eraser, and fire hood (from left to right).}
  \label{fig:full_vis_2}
\end{figure*}

\begin{figure*}[h]
  \centering
  \begin{subfigure}{\linewidth}
    \centering
    \includegraphics[width=\linewidth]{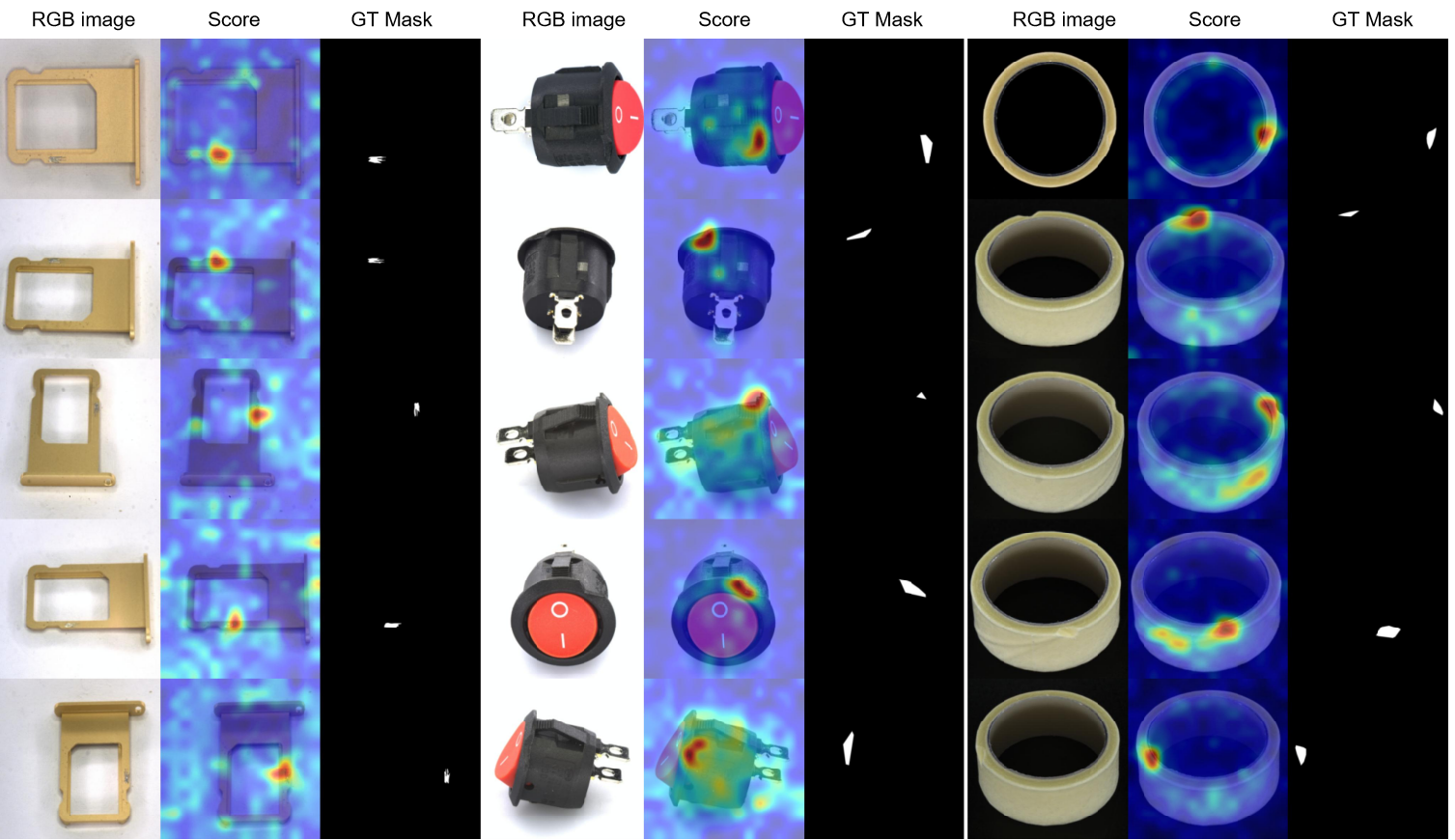}
    \label{fig:short-a}
  \end{subfigure}
  \caption{Visualization of our method on sim card set, switch and tape (from left to right).}
  \label{fig:full_vis_3}
\end{figure*}

\begin{figure*}[h]
  \centering
  \begin{subfigure}{\linewidth}
    \centering
    \includegraphics[width=\linewidth]{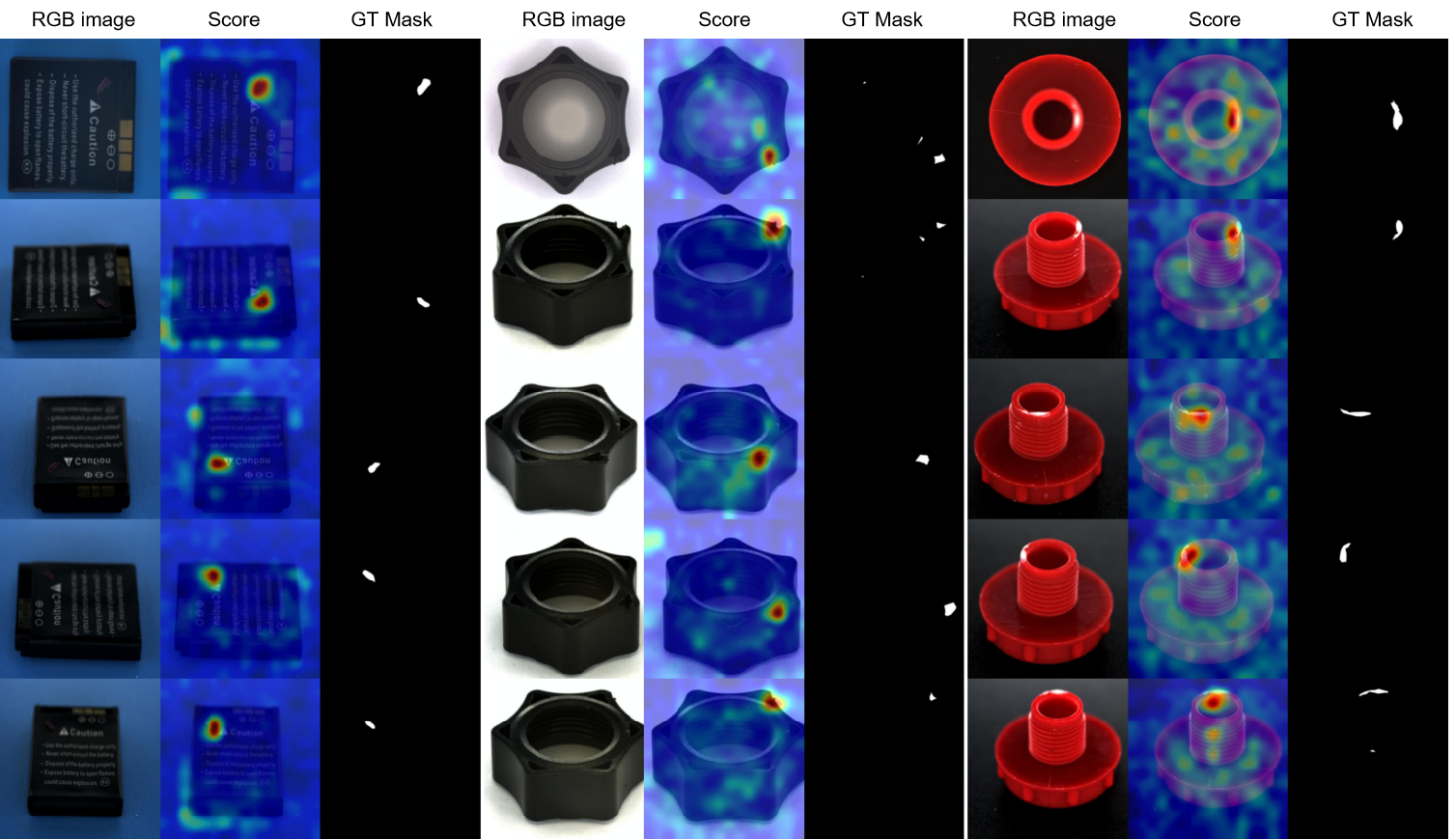}
    \label{fig:short-a}
  \end{subfigure}
  \caption{Visualization of our method on phone battery, plastic nut and plastic plug (from left to right).}
  \label{fig:full_vis_4}
\end{figure*}

\begin{figure*}[h]
  \centering
  \begin{subfigure}{\linewidth}
    \centering
    \includegraphics[width=\linewidth]{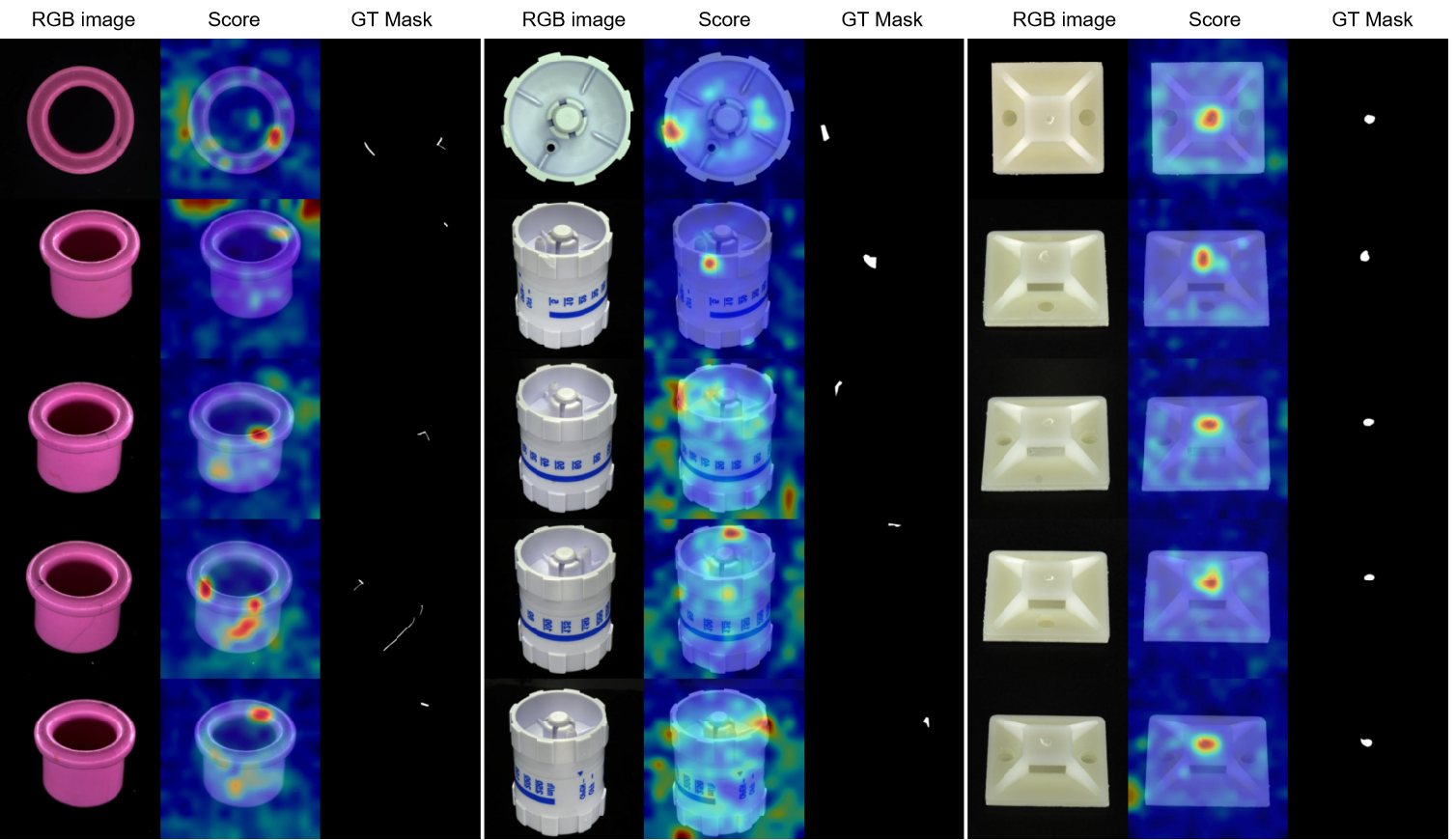}
    \label{fig:short-a}
  \end{subfigure}
  \caption{Visualization of our method on porcelain doll, regulator and rolled strip base (from left to right).}
  \label{fig:full_vis_5}
\end{figure*}

\begin{figure*}[h]
  \centering
  \begin{subfigure}{\linewidth}
    \centering
    \includegraphics[width=\linewidth]{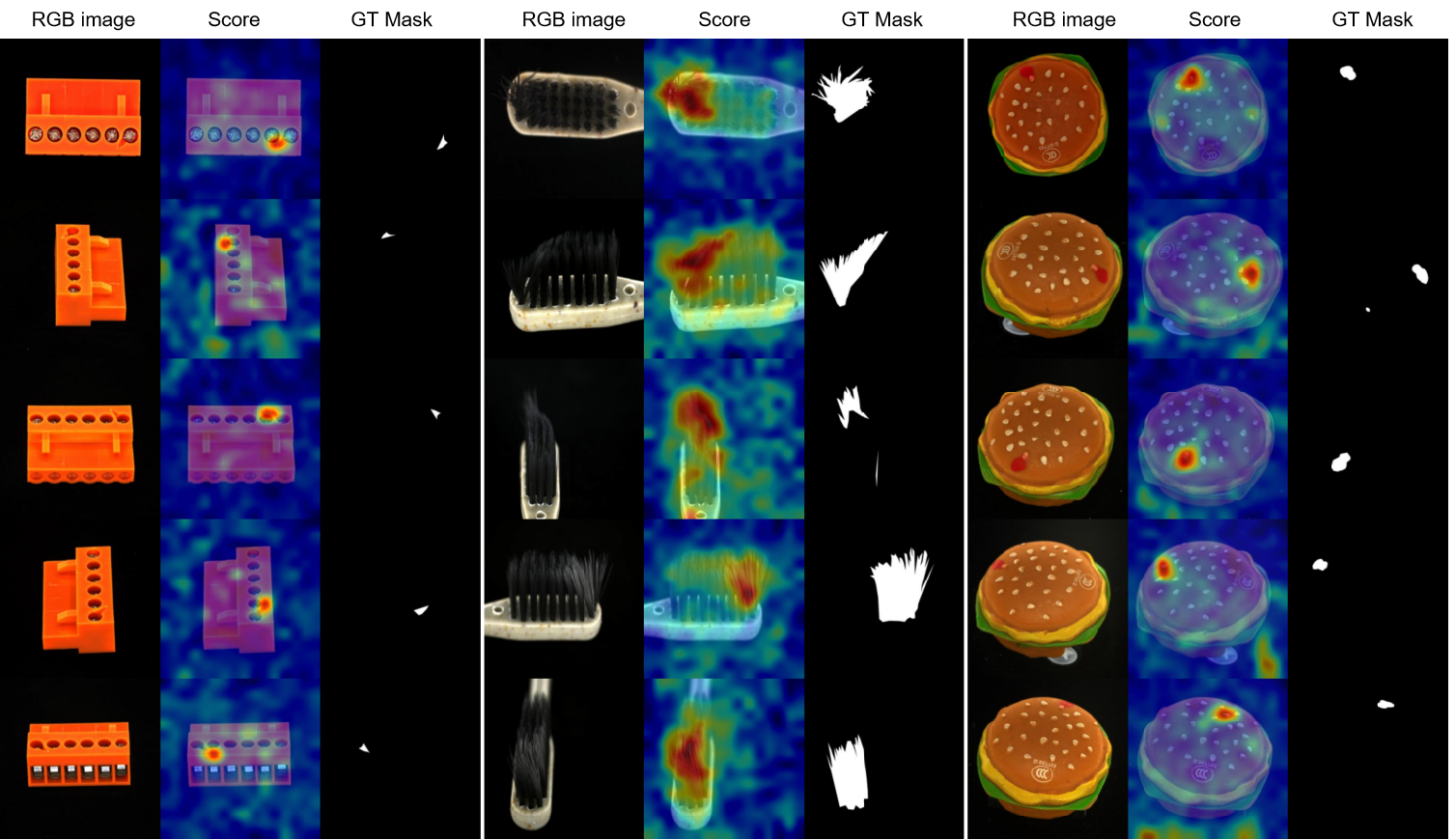}
    \label{fig:short-a}
  \end{subfigure}
  \caption{Visualization of our method on terminalblock, toothbrush and toy (from left to right).}
  \label{fig:full_vis_7}
\end{figure*}

\begin{figure*}[h]
  \centering
  \begin{subfigure}{\linewidth}
    \centering
    \includegraphics[width=\linewidth]{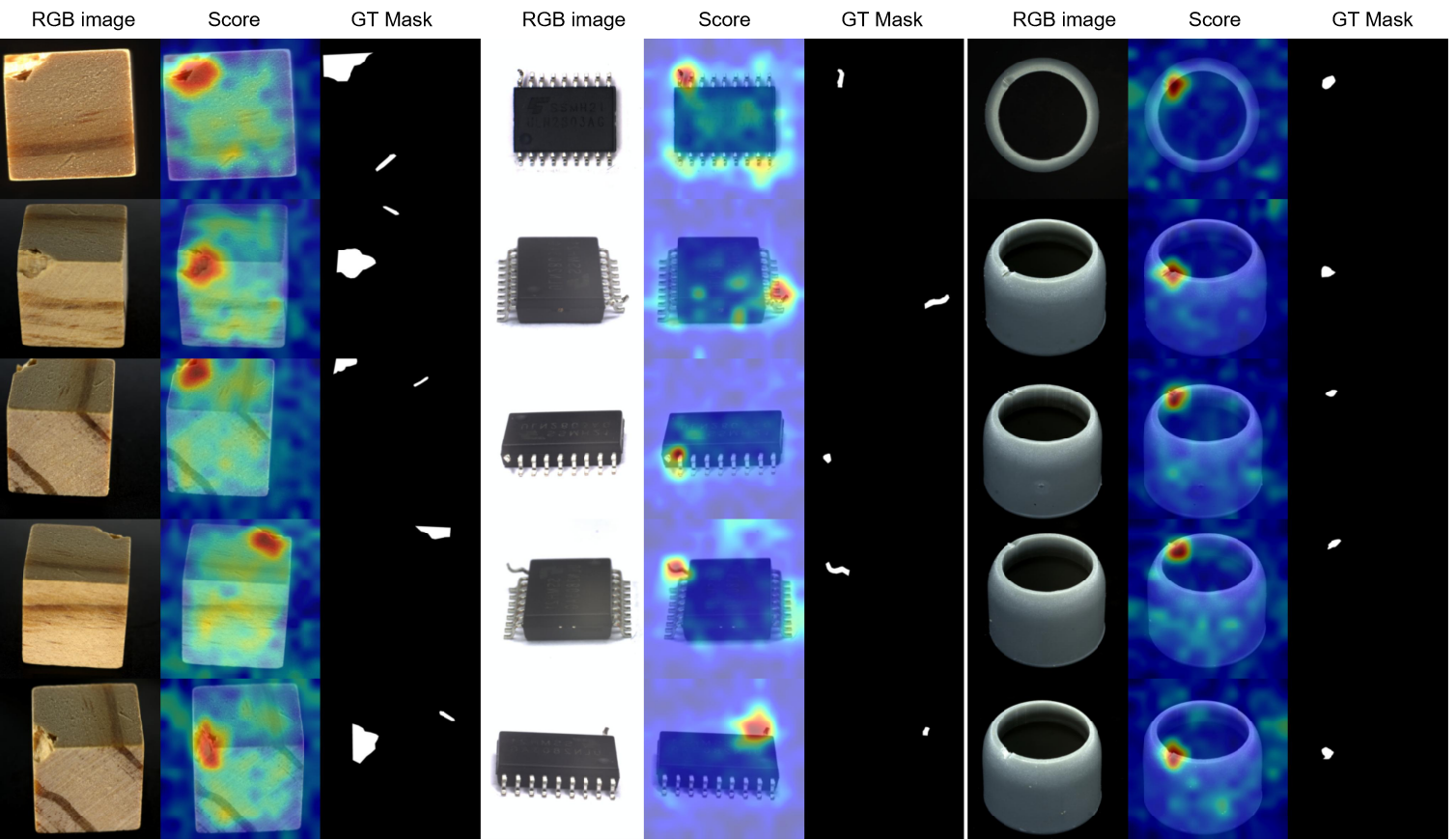}
    \label{fig:short-a}
  \end{subfigure}
  \caption{Visualization of our method on toy brick, transistor1 and u block (from left to right).}
  \label{fig:full_vis_8}
\end{figure*}

\begin{figure*}[h]
  \centering
  \begin{subfigure}{\linewidth}
    \centering
    \includegraphics[width=\linewidth]{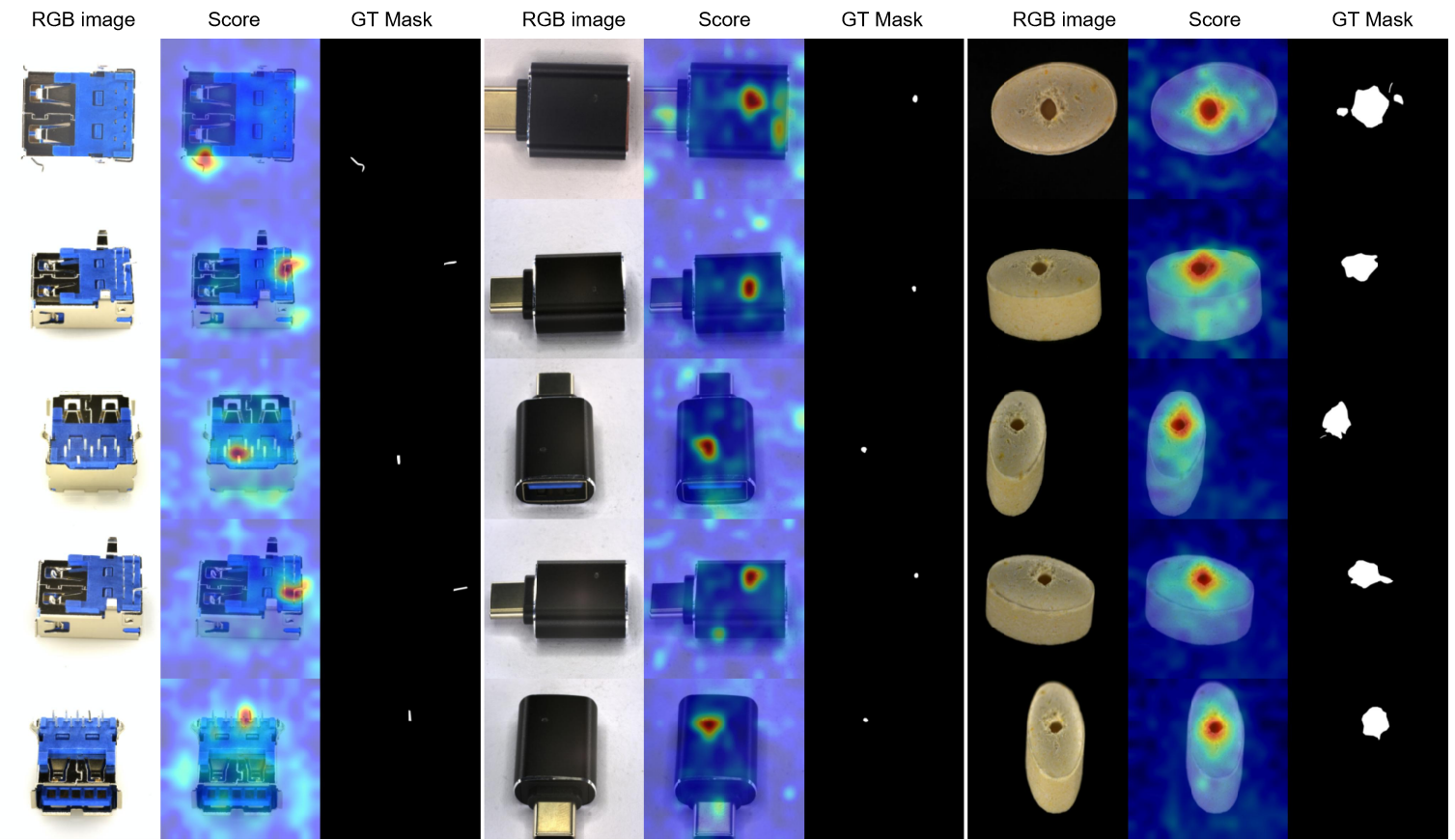}
    \label{fig:short-a}
  \end{subfigure}
  \caption{Visualization of our method on usb, usb adaptor and vcpill (from left to right).}
  \label{fig:full_vis_9}
\end{figure*}

\begin{figure*}[h]
  \centering
  \begin{subfigure}{\linewidth}
    \centering
    \includegraphics[width=\linewidth]{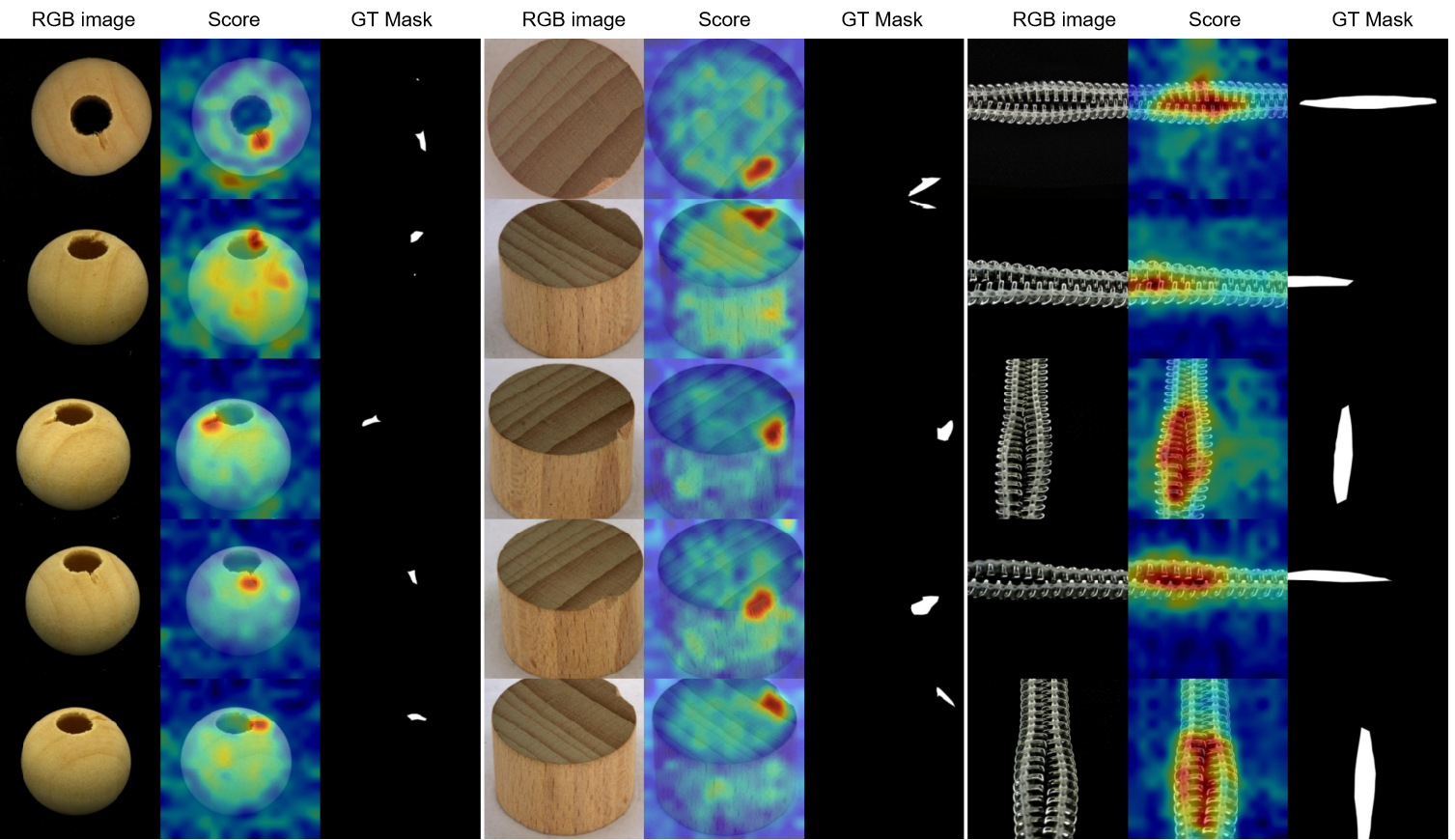}
    \label{fig:short-a}
  \end{subfigure}
  \caption{Visualization of our method on wooden beads, woodstick and zipper (from left to right).}
  \label{fig:full_vis_10}
\end{figure*}

\Cref{fig:full_vis_1,fig:full_vis_2,fig:full_vis_3,fig:full_vis_4,fig:full_vis_5,fig:full_vis_6,fig:full_vis_7,fig:full_vis_8,fig:full_vis_9,fig:full_vis_10} presents the visualization results of our method on 30 subclasses of the Real-IAD dataset, with one sample per subclass.

\section*{D. Pseudocode for Multi-center Pre-training}

\cref{alg:eam_pretraining} provides the algorithmic pseudocode for multi-center pre-training strategy discussed in Sec. (3.3).

\begin{algorithm}[htbp]
\caption{Pre-training Procedure for Epipolar Attention Module (EAM)}
\label{alg:eam_pretraining}
\begin{algpseudocode}
\textbf{Input:} Dataset $\mathcal{D}$, Frozen Backbone $f(\theta)$, Fundamental Matrices $F \in \mathbb{R}^{(V-1) \times (V-1) \times 3 \times 3}$, Random Initialized EAM $g_0(F, \theta)$, Number of epochs $N$

\textbf{Output:} Trained EAM $g(F, \theta)$

\For {$epoch = 1$ \textbf{to} $N$}

    \For {each batch $X_i \in \mathcal{D}$}
        \State $Z_i \leftarrow f(X_i; \theta)$
        \State $r \sim \text{Uniform}(\{0,1,\dots,V-1\})$
        \State $Z_i^{ref} \leftarrow Z_i[r]$, $Z_i^{ref-} \leftarrow \text{RandomMask}(Z_i^{ref})$ 
        \State $Z_i^{-} \leftarrow (Z_i - Z_i^{ref}) \cup Z_i^{ref-}$ 
        \State $\tilde{Z}_i \leftarrow g(Z_i, F)$, $\tilde{Z}_i^{-} \leftarrow g(Z_i^{-}, F)$
        \State Compute $p_{neg}$ via Eq. (8)
        \State Compute loss $\mathcal{L}$ via Eq. (10)
        \State Update $g(F,\theta)$ by optimizing $\mathcal{L}$
    \EndFor
    
\EndFor

\Return $g(F,\theta)$
\end{algpseudocode}
\end{algorithm}

\begin{figure*}[h]
  \centering
  \begin{subfigure}{\linewidth}
    \centering
    \includegraphics[width=\linewidth]{ICCV_Multiview_Anomalydetection/sec/figure/full_vis/sim_card_set_switch_tape.png}
    \label{fig:short-a}
  \end{subfigure}
  \caption{Visualization of our method on sim card set, switch and tape (from left to right).}
  \label{fig:full_vis_6}
\end{figure*}

\section*{E. Pixel-Level Evaluation}

\cref{tab:pixel_full} presents the pixel-level AUROC results under both single-class and multi-class settings. Our method achieves state-of-the-art (SOTA) performance in the multi-class setting, demonstrating its effectiveness in fine-grained anomaly localization.

Furthermore, we validate that replacing WideResNet50, which excels at distinguishing fine-grained structures, with DINOv2 does not lead to a significant decline in segmentation performance. This finding highlights that our approach remains highly competitive at the pixel level.

\section*{F. Broader Impact and Limitations}

\subsection*{F.1 Broader Impact}

The proposed method could improve the efficacy of industrial anomaly detection with multi-camera system. The method utilizes epipolar geometry to constrain cross-view attention thus improving the effectiveness for both view independent and sample level anomaly detection. Additionally, the method achieves a level of efficiency sufficient for real-time detection. Potential negative impacts include reducing reliance on human operators for quality inspection. 

\subsection*{F.2 Potential Limitations}

The proposed method requires substantial camera view overlap between different viewpoints. If there is no overlap between the field-of-view of different cameras, this method will degenerate into regular single-view anomaly detection method.

\section*{E. Public Resource Used}
\label{sec:ack}
In this section, we acknowledge the use of the following public resources, during this work:

\begin{itemize}
    \item Pytorch \footnote{\url{https://github.com/pytorch/pytorch}} \dotfill Pytorch License 
    \item Real-IAD \footnote{\url{https://github.com/Tencent/AnomalyDetection_Real-IAD}} \dotfill Apache License 2.0 
    \item Transformers \footnote{\url{https://github.com/huggingface/transformers}} \dotfill Apache License 2.0 
    \item ADer(UniAD) \footnote{\url{https://github.com/zhangzjn/ADer}} \dotfill MIT License
    \item PatchCore \footnote{\url{https://github.com/amazon-science/patchcore-inspection}} \dotfill Apache License 2.0 
    
\end{itemize}

\begin{table*}[t]
  \centering
  \caption{Comparison of Pixel-level AUROC results for our method on Real-IAD dataset with single-class/multi-class setting.}
  \resizebox{0.8\linewidth}{!}{
    \begin{tabular}{lccccc}
      \toprule
      \multicolumn{1}{c}{} & \multicolumn{5}{c}{AUROC}  \\
      \cmidrule(lr){2-6} \cmidrule(lr){7-11}
      Category & UniAD & SimpleNet & MVAD & PatchCore & Ours\\
            \midrule
     audiojack & 97.2/97.6 & 98.2/74.4 & 98.8/97.1 & 98.4/\textbf{98.1} & \textbf{99}/96.5\\
    bottle\_cap & 99.2/99.5 & 98.5/85.3 & \textbf{99.7}/\textbf{99.6} & 99.2/96.6 & 99.5/99.5\\
    button\_battery & 93.7/96.7 & 98.0/75.9 & \textbf{98.8}/97.9 & 98.2/98.2 & \textbf{98.8}/\textbf{98.3}\\
    end\_cap & 96.7/95.8 & 94.2/63.1 & \textbf{98.1}/\textbf{96.7} & 96.2/96.4 & 98.0/96.5\\
    eraser & 99.0/\textbf{99.3} & 98.3/80.6 & \textbf{99.2}/99.2 & 98.6/98.3 & \textbf{99.2}/99.2\\
    fire\_hood & 98.5/98.6 & 97.5/70.5 & \textbf{99.1}/98.7 & 98.4/\textbf{98.8} & \textbf{99.1}/98.6\\
    mint & 94.3/94.4 & 94.1/79.9 & \textbf{98.5}/\textbf{95.9} & 95.4/95.4 & 98.1/95.6\\
    mounts & \textbf{99.4}/\textbf{99.4} & 98.0/80.5 & 99.0/99.2 & 97.6/96.4 & 98.8/96.2\\
    pcb & 96.6/97.0 & 98.4/78.0 & \textbf{99.4}/\textbf{98.8} & 97.2/96.2 & 98.7/96.3\\
    phone\_battery & 97.9/85.5 & 96.5/43.4 & \textbf{99.2}/80.8 & 99.1/95.9 & 99.1/\textbf{97.6}\\
    plastic\_nut & 98.6/98.4 & 98.1/77.4 & \textbf{99.6}/\textbf{99.2} & 98.4/97.7 & 99.3/98.9\\
    plastic\_plug & 97.9/98.6 & 96.1/78.6 & \textbf{99.0}/\textbf{99.1} & 96.6/95.7 & 98.8/99.0\\
    porcelain\_doll & 97.3/98.7 & 96.6/81.8 & \textbf{99.1}/99.2 & 95.6/92.8 & 98.5/\textbf{99.3}\\
    regulator & 93.7/95.5 & 97.0/76.6 & \textbf{99.1}/97.1 & 95.2/91.1 & 97.2/\textbf{98.3}\\
    rolled\_strip\_base & 98.9/\textbf{99.6} & 98.8/80.5 & \textbf{99.7}/99.5 & 99.5/99.2 & 99.6/98.8\\
    sim\_card\_set & 96.7/97.9 & 97.3/71.0 & 98.5/\textbf{98.9} & 98.8/98.7 & \textbf{98.9}/98.6\\
    switch & 99.4/98.1 & 99.1/71.7 & 99.5/\textbf{98.7} & 99.2/98.5 & \textbf{99.6}/97.4\\
    tape & 99.5/\textbf{99.7} & 99.2/77.5 & \textbf{99.7}/\textbf{99.7} & \textbf{99.7}/99.1 & \textbf{99.7}/99.5\\
    terminalblock & 98.9/99.2 & 99.3/87.0 & \textbf{99.8}/\textbf{99.6} & 98.9/96.9 & 99.4/96.4\\
    toothbrush & 96.8/95.7 & 94.3/84.7 & 97.3/97.1 & 94.5/95.0 & \textbf{98.3}/\textbf{97.3}\\
    toy & 96.4/93.4 & 91.9/67.7 & \textbf{97.3}/\textbf{95.9} & 91.8/91.8 & 96.0/94.4\\
    toy\_brick & \textbf{97.9}/\textbf{97.4} & 94.3/86.5 & 97.6/96.0 & 95.5/97.0 & 96.9/96.5\\
    transistor1 & 98.8/98.9 & 99.1/71.7 & \textbf{99.5}/\textbf{99.3} & 98.9/98.7 & 99.2/98.1\\
    u\_block & 99.0/99.3 & 98.6/76.2 & \textbf{99.6}/\textbf{99.5} & 99.3/98.6 & 99.3/99.3\\
    usb & 98.5/97.9 & 98.9/81.1 & \textbf{99.6}/99.2 & 98.6/98.1 & 99.2/\textbf{99.3}\\
    usb\_adaptor & 97.0/96.6 & 95.7/67.9 & \textbf{97.3}/97.1 & 93.7/92.3 & 96.6/\textbf{98.1}\\
    vcpill & \textbf{99.1}/\textbf{99.1} & 98.6/68.2 & 99.0/98.6 & 98.5/98.6 & 98.4/97.4\\
    wooden\_beads & 97.5/97.6 & 96.7/68.1 & \textbf{98.6}/98.1 & \textbf{98.6}/\textbf{98.5} & \textbf{98.6}/97.9\\
    woodstick & 96.6/94.0 & 93.5/76.1 & 98.5/97.4 & 98.3/98.6 & \textbf{98.6}/\textbf{98.7}\\
    zipper & 97.5/98.4 & 98.6/89.9 & \textbf{99.2}/\textbf{99.1} & 98.9/98.9 & 99.0/98.0\\
    \midrule
    Average & 97.6/97.3 & 96.8/75.7 & \textbf{98.9}/97.7 & 97.6/96.9 & 98.6/\textbf{97.9}\\

      \bottomrule
    \end{tabular}
  }
  \label{tab:pixel_full}
\end{table*}

{
    \small
    \bibliographystyle{ieeenat_fullname}
    \bibliography{main}
}